\documentclass{article}

\PassOptionsToPackage{numbers, compress}{natbib}

\usepackage[final]{neurips_2022}

\usepackage[utf8]{inputenc} 
\usepackage[T1]{fontenc}    
\usepackage{hyperref}       
\usepackage{url}            
\usepackage{booktabs}       
\usepackage{amsfonts}       
\usepackage{nicefrac}       
\usepackage{microtype}      
\usepackage{xcolor}         

\usepackage{custom}
\usepackage[normalem]{ulem}

\usetikzlibrary{external}
\title{Rethinking Value Function Learning for Generalization in Reinforcement Learning}

\author{
  Seungyong Moon$^{1, 2}$, JunYeong Lee$^{1, 2}$, Hyun Oh Song$^{1,2,3}$\thanks{Corresponding author} \\
  $^1$Seoul National University, $^2$Neural Processing Research Center, $^3$DeepMetrics \\
  \texttt{\{symoon11,mascheroni99,hyunoh\}@mllab.snu.ac.kr} \\
}

\begin{document}

\maketitle

\begin{abstract}
Our work focuses on training RL agents on multiple visually diverse environments to improve observational generalization performance. In prior methods, policy and value networks are separately optimized using a disjoint network architecture to avoid interference and obtain a more accurate value function. We identify that a value network in the multi-environment setting is more challenging to optimize and prone to memorizing the training data than in the conventional single-environment setting. In addition, we find that appropriate regularization on the value network is necessary to improve both training and test performance. To this end, we propose Delayed-Critic Policy Gradient (DCPG), a policy gradient algorithm that implicitly penalizes value estimates by optimizing the value network less frequently with more training data than the policy network. This can be implemented using a single unified network architecture. Furthermore, we introduce a simple self-supervised task that learns the forward and inverse dynamics of environments using a single discriminator, which can be jointly optimized with the value network. Our proposed algorithms significantly improve observational generalization performance and sample efficiency on the Procgen Benchmark.
\end{abstract}

\section{Introduction}

In recent years, deep reinforcement learning (RL) has achieved remarkable success in various domains, such as robotic controls and games \citep{mnih2015human,haarnoja2018soft,schrittwieser2020mastering}. To apply RL algorithms to more practical scenarios, such as autonomous vehicles or healthcare systems, they should be robust against the non-stationarity of real-world environments and capable of performing well on unseen situations during deployment. However, current state-of-the-art RL algorithms often fail to generalize to unseen test environments with visual variations, even if they achieve high performance in their training environments \citep{farebrother2018generalization,zhang2018study,cobbe2019quantifying}. This problem is referred to as \emph{observational overfitting} \citep{song2020observational}.

Training RL agents on a finite number of visually diverse environments and testing them on unseen environments is the standard protocol for evaluating observational generalization in RL \cite{cobbe2020leveraging}. Several methods have attempted to improve generalization in this framework by adopting the regularization techniques that originate from supervised learning or training robust state representations via self-supervised learning \cite{cobbe2019quantifying,igl2019generalization,raileanu2021automatic,mazoure2022crosstrajectory}. However, these methods have mainly focused on developing new auxiliary objectives on the existing RL algorithms intended for the conventional single-environment setting such as PPO \cite{schulman2017proximal}. Some recent works have investigated the interference between policy and value function optimization arising from the multiple training environments and proposed new training schemes that decouple the policy and value network training with a separate network architecture to obtain an accurate value function \cite{cobbe2021phasic,raileanu2021decoupling}.

In this paper, we argue that learning an accurate value function on multiple training environments is more challenging than on a single training environment and requires sufficient regularization. We demonstrate that a value network trained on multiple environments is more likely to memorize the training data and cannot generalize to unvisited states within the training environments, which can be detrimental to not only training performance but also test performance on unseen environments. In addition, we find that regularization techniques that penalize large estimates of the value network, originally developed for preventing memorization in the single-environment setting, are also beneficial for improving both training and test performance in the multi-environment setting. However, this benefit comes at the cost of premature convergence, which hinders further performance enhancement.

To address this, we propose a new model-free policy gradient algorithm named \emph{Delayed-Critic Policy Gradient} (DCPG), which trains the value network with lower update frequency but with more training data than the policy network. We find that the value network with delayed updates suffers less from the memorization problem and significantly improves training and test performance. In addition, we demonstrate that it provides better state representations to the policy network using a single unified network architecture, unlike the prior methods. Moreover, we introduce a simple self-supervised task that learns the forward and inverse dynamics of environments using a single discriminator on top of DCPG. Our algorithms achieve state-of-the-art observational generalization performance and sample efficiency compared to prior model-free methods on the Procgen benchmark \cite{cobbe2020leveraging}.

\section{Preliminaries}

\subsection{Observational Generalization in RL}

We consider a collection of environments $\mathcal{M}$ formulated as Markov Decision Processes (MDPs). Each environment $m \in \mathcal{M}$ is described as a tuple $(\mathcal{S}_m, \mathcal{A}, T_m, r_m, \rho_m, \gamma)$, where $\mathcal{S}_m$ is the image-based state space, $\mathcal{A}$ is the action space shared across all environments, $T_m: \mathcal{S}_m \times \mathcal{A} \to \mathcal{P}(\mathcal{S}_m)$ is the transition function, $r_m: \mathcal{S}_m \times \mathcal{A} \to \mathbb{R}$ is the reward function, $\rho_m$ is the initial state distribution, and $\gamma \in [0, 1]$ is the discount factor. We assume that the state space has visual variations between different environments. While the transition and reward functions are defined as specific to an environment, we assume that they exhibit some common structures across all environments. A policy $\pi: \mathcal{S} \to \mathcal{P}(\mathcal{A})$ is trained on a finite number of training environments $\mathcal{M}_\mathrm{train} = \{m_i\}_{i=1}^{n}$, where $\mathcal{S}$ is the set of all possible states in $\mathcal{M}$. Our goal is to learn a generalizable policy that maximizes the expected return on unseen test environments $\mathcal{M}_\mathrm{test} = \mathcal{M} \setminus \mathcal{M}_\mathrm{train}$.

In this paper, we utilize the Procgen benchmark as a testbed for observational generalization \citep{cobbe2020leveraging}. It is a collection of 16 video games with high diversity comparable to the ALE benchmark \cite{bellemare2013arcade}. Each game consists of procedurally generated environment instances with visually different layouts, backgrounds, and game entities (\eg, the spawn locations and times for enemies), also called levels. The standard evaluation protocol on the Procgen benchmark is to train a policy on a finite set of training levels and evaluate its performance on held-out test levels \cite{cobbe2020leveraging}.

\subsection{Proximal Policy Optimization}

Proximal Policy Optimization (PPO) is a powerful model-free policy gradient algorithm that learns a policy $\pi_\theta$ and value function $V_\phi$ parameterized by deep neural networks \citep{schulman2017proximal}. For training, PPO first collects trajectories $\tau$ using the old policy network $\pi_{\theta_\mathrm{old}}$ right before the update. Then, the policy network is trained with the collected trajectories for several epochs to maximize the following clipped surrogate policy objective $J_\pi$ designed to constrain the size of policy update:
\begin{align*}
    J_\pi(\theta) = \mathbb{E}_{s_t, a_t \sim \tau} \left[ \min \left( \frac{\pi_\theta(a_t \mid s_t)}{\pi_{\theta_\mathrm{old}}(a_t \mid s_t)} \hat{A}_t, \ \mathrm{clip}\left( \frac{\pi_\theta(a_t \mid s_t)}{\pi_{\theta_\mathrm{old}}(a_t \mid s_t)}, 1 - \epsilon, 1 + \epsilon \right) \hat{A}_t \right) \right],
\end{align*}
where $\hat{A}_t$ is an estimate of the advantage function at timestep $t$. Concurrently, the value network is trained with the collected trajectories to minimize the following value objective $J_V$:
\begin{align*}
    J_V(\phi) = \mathbb{E}_{s_t \sim \tau} \left[ \frac{1}{2} \left( V_\phi(s_t) - \hat{R}_t \right)^2 \right],
\end{align*}
where $\hat{R}_t = \hat{A}_t + V_\phi(s_t)$ is the value function target. It is used to compute the advantage estimates via generalized advantage estimator (GAE) \citep{schulman2016high}.

In practice, the policy and value networks are jointly optimized with shared parameters (\ie, $\theta=\phi$), especially in image-based RL \citep{baselines,yarats2019improving}. For example, they can be implemented using a shared encoder followed by separate linear heads, as shown in \Cref{fig:network_ppo}. Sharing parameters is advantageous in that representations learned by each objective can be beneficial to the other. It also reduces memory costs and accelerates training time. However, a shared network architecture complicates the optimization as a single encoder should be optimized over multiple objectives whose gradients may have varying scales and directions. It also constrains the policy and value networks to be optimized under the same training hyperparameter setting, such as batch size and the number of epochs, severely limiting the flexibility of PPO.

\newcommand{\nodecaption}[2]{\subcaptionbox{#1\label{#2}}{\phantom{(X)~#1}}}

\begin{figure}[t]
    \centering
    \begin{tikzpicture}
    	\node[rectangle,draw,minimum height=1.6cm,minimum width=1.6cm,align=center,line width=0.2mm,font=\small] (enc) at (0,0) {Encoder \\ $\theta$};
    	\node[above left=0.6cm and 0.7cm of enc.north,align=center,anchor=south,font=\small] (pi) {Policy \\ $\pi_\theta (\cdot \mid s)$};
    	\draw [-stealth,line width=0.2mm](enc.north) -- (pi.south);
    	\node[above=0.1cm of pi.north,align=center,font=\small] (pi_obj) {$\blue{J_\pi(\theta)}$ \\[0.3em] \white{$C_\phi(\theta)$}};
	\node[above left=0.0cm and 0.1cm of pi_obj.west,anchor=east,align=right,font=\small] (phase) {Policy phase \\[0.3em] Aux phase};
    	\node[above right=0.6cm and 0.7cm of enc.north,align=center,anchor=south,font=\small] (v) {Value \\ $V_\theta (s)$};
    	\draw [-stealth,line width=0.2mm](enc.north) -- (v.south);
    	\node[above=0.1cm of v.north,align=center,font=\small] (v_obj) {$\blue{J_V(\theta)}$ \\[0.3em] \white{$C_V(\theta)$} };
    	\node[below=0.6cm of enc.south,font=\small] (s) {$s$};
    	\draw [-stealth,line width=0.2mm](s.north) -- (enc.south);
	\node (c) at (-0.1,-2.0)  {\nodecaption{PPO network}{fig:network_ppo}};
	\draw [densely dashed,line width=0.2mm](1.4,-1.7) -- (1.4,3.4);
	\node[rectangle,draw,minimum height=1.6cm,minimum width=1.6cm,align=center,line width=0.2mm,font=\small] (enc) at (3.0,0) {Policy \\ encoder \\ $\theta$};
	\node[above left=0.6cm and 0.7cm of enc.north,align=center,anchor=south,font=\small] (pi) {Policy \\ $\pi_\theta (\cdot \mid s)$};
    	\draw [-stealth,line width=0.2mm](enc.north) -- (pi.south);
    	\node[above=0.1cm of pi.north,align=center,font=\small] (pi_obj) {$\blue{J_\pi(\theta)}$ \\[0.3em] $\red{C_\pi(\theta)}$};
    	\node[above right=0.6cm and 0.7cm of enc.north,align=center,anchor=south,font=\small] (aux_v) {Aux value \\ $V_\theta (s)$};
    	\draw [-stealth,line width=0.2mm,font=\small](enc.north) -- (aux_v.south);
    	\node[above=0.1cm of aux_v.north,align=center,font=\small] (aux_v_obj) {$\red{J_{\mathrm{aux}}(\theta)}$};
    	\node[below=0.6cm of enc.south,font=\small] (s) {$s$};
    	\draw [-stealth,line width=0.2mm](s.north) -- (enc.south);
	\node[rectangle,draw,minimum height=1.6cm,minimum width=1.6cm,align=center,line width=0.2mm,font=\small] (enc) at (5.2,0) {Value \\ encoder \\ $\phi$};
    	\node[above=0.6cm of enc.north,align=center,font=\small] (v) {Value \\ $V_\phi (s)$};
    	\draw [-stealth,line width=0.2mm,font=\small](enc.north) -- (v.south);
    	\node[above=0.1cm of v.north,align=center,font=\small] (v_obj) {$\blue{J_V(\phi)}$ \\[0.3em] $\red{J_V(\phi)}$};
    	\node[below=0.6cm of enc.south,font=\small] (s) {$s$};
    	\draw [-stealth,line width=0.2mm](s.north) -- (enc.south);
	\node (c) at (3.9,-2.0) {\nodecaption{PPG network}{fig:network_ppg}};
	\draw [densely dashed,line width=0.2mm](6.4,-1.7) -- (6.4,3.4);
	\node[rectangle,draw,minimum height=1.6cm,minimum width=1.6cm,align=center,line width=0.2mm,font=\small] (enc) at (8.7,0) {Encoder \\ $\theta$};
	\node[above left=0.6cm and 1.4cm of enc.north,align=center,anchor=south,font=\small] (pi) {Policy \\ $\pi_\theta (\cdot \mid s)$};
	\draw [-stealth,line width=0.2mm](enc.north) -- (pi.south);
	\node[above=0.1cm of pi.north,align=center,font=\small] (pi_obj) {$\blue{J_\pi(\theta)}$ \\[0.3em] $\red{C_\pi(\theta)}$};
	\node[above=0.6cm of enc.north,align=center,anchor=south,font=\small] (v) {Value \\ $V_\theta (s)$};
	\draw [-stealth,line width=0.2mm](enc.north) -- (v.south);
	\node[above=0.1cm of v.north,align=center,font=\small] (v_obj) {$\blue{C_V(\theta)}$ \\[0.3em] $\red{J_V(\theta)}$};
	\node[above right=0.6cm and 1.4cm of enc.north,align=center,anchor=south,font=\small] (f) {\cyan{Dynamics} \\ \cyan{$f_\theta(s, a, s')$}};
	\draw [-stealth,line width=0.2mm](enc.north) -- (f.south);
	\node[above=0.1cm of f.north,align=center,font=\small] (f_obj) {$\white{C_V(\theta)}$ \\[0.3em] $\red{J_f(\theta)}$};
	\node[below=0.6cm of enc.south,font=\small] (s) {$s, \cyan{s'}$};
	\draw [-stealth,line width=0.2mm](s.north) -- (enc.south);	
	\begin{scope}[transform canvas={xshift=0.2em}]
		\node[below=0.6cm of f.south,align=center,font=\small] (a) {$\cyan{a}$};
		\draw [-stealth,line width=0.2mm](a.north) -- (f.south);
	\end{scope}
	\node (c) at (8.6,-2.0) {\nodecaption{\cyan{D}DCPG network}{fig:network_dcpg}};
    \end{tikzpicture}
    \caption{Network architectures for PPO, PPG, and \cyan{D}DCPG. The objectives $J_\pi$, $J_V$, $J_\mathrm{aux}$, and $J_f$ denote the policy, value, auxiliary value, and dynamics objectives, respectively. The regularizers $C_\pi$ and $C_V$ denote the policy and value regularizers, respectively. The blue and red terms represent optimization problems during the policy and auxiliary phases, respectively.}
    \label{fig:network}
\end{figure}
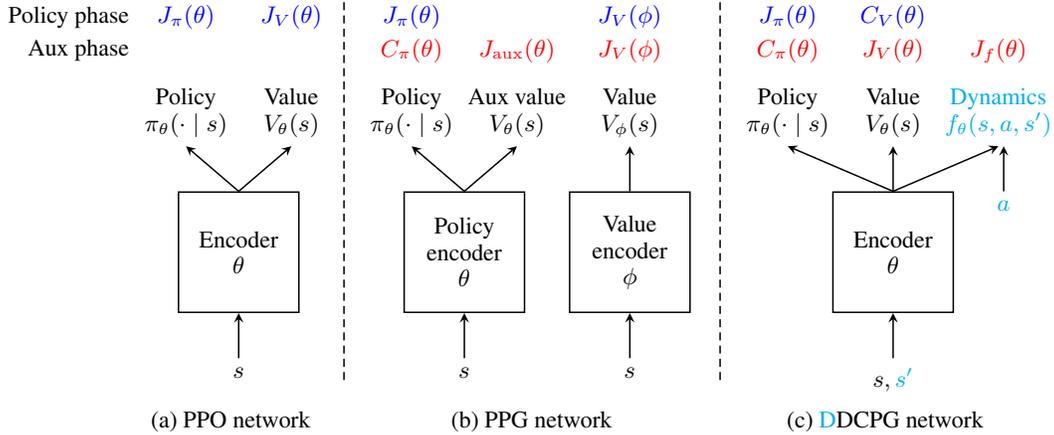

\subsection{Phasic Policy Gradient}

Phasic Policy Gradient (PPG) is an algorithm built upon PPO that significantly improves observational generalization by addressing the problems of sharing parameters \citep{cobbe2021phasic}. More specifically, PPG employs separate encoders for the policy and value networks, as shown in \Cref{fig:network_ppg}. In addition, it introduces an auxiliary value head $V_\theta$ on top of the policy encoder in order to distill useful representations from the value network into the encoder. For training, PPG alternates between policy and auxiliary phases. During the policy phase, which is repeated $N_\pi$ times, the policy and value networks are trained with newly-collected trajectories to optimize the policy and value objectives from PPO, respectively. Then, all states and value function targets in the trajectories are stored in a buffer $\mathcal{B}$. During the auxiliary phase, the auxiliary value head and the policy network are jointly trained with all data in the buffer to optimize the following auxiliary value objective $J_\mathrm{aux}$ and policy regularizer $C_\pi$:
\begin{align*}
	J_\mathrm{aux}(\theta) = \mathbb{E}_{s_t \sim \mathcal{B}} \left[ \frac{1}{2} \left( V_\theta(s_t) - \hat{R}_t \right)^2 \right], \ C_\pi(\theta) = \mathbb{E}_{s_t \sim \mathcal{B}} \left[ D_\mathrm{KL}(\pi_{\theta_\mathrm{old}}(\cdot \mid s_t) \parallel \pi_\theta(\cdot \mid s_t))\right],
\end{align*}
where $\pi_{\theta_\mathrm{old}}$ is the policy network right before the auxiliary phase and $D_\mathrm{KL}$ denotes the KL divergence. In other words, the value network is distilled into the policy encoder while maintaining the outputs of the policy network unchanged. Moreover, the value network is additionally trained with all data in the buffer to optimize the value objective from PPO to obtain a more accurate value function. It is worth noting that the training data size in the auxiliary phase is $N_\pi$ times larger than the policy phase. It has been claimed that the distillation of a better-trained value network with a separate architecture and the additional training for a more accurate value network can improve observational generalization performance and sample efficiency \citep{cobbe2021phasic}.

\section{Motivation}
\label{sec:motivation}

\subsection{Difficulty of Training Value Network on Multiple Training Environments}
\label{subsec:memorization}

We begin by investigating the difficulty of obtaining an accurate value network across multiple training environments. Indeed, learning a value network that better approximates the true value function on the given training environments can result in improved training performance \cite{sutton1999policy}. However, even in a simple setting where an agent is trained on a single environment, it has been shown that a value network is likely to memorize the training data and unable to extrapolate well to unseen states even in the same training environment \citep{kolter2009regularization,ilyas2020a,dabney2020value,flet-berliac2021learning}. This problem can be exacerbated when the number of training environments increases. Intuitively, given the fixed number of environment steps, the value network will be provided fewer training samples per environment and rely more on memorization.

To corroborate this claim, we measure the stiffness of the value network between states $(s, s')$ while varying the number of training environments \citep{fort2019stiffness,bengio2020interference}, which is defined by
\begin{align*}
	\rho(s, s') = \frac{\nabla_\phi J_V(\phi; s)^\intercal \nabla_\phi J_V(\phi; s')}{\| \nabla_\phi J_V(\phi; s) \|_2 \|\nabla_\phi J_V(\phi; s') \|_2}.
\end{align*}
Low stiffness indicates that updating the network parameters toward minimizing the value objective for one state will have a negative effect on the minimization of the value objective for other states \citep{bengio2020interference}. That is, the value network is less able to adjust its parameter to predict the true value function across different states and instead tends to memorize only the states it has encountered. More specifically, we train PPG agents on the Procgen games while increasing the number of training levels from 0 to 200 and compute the average stiffness across all state pairs in a mini-batch of size $2^{14}$ (=16,384) throughout training. The detailed experimental settings and results can be found in Appendix A.

The green lines in \Cref{fig:cossim_value_selective_both} show that the stiffness of the value network decreases as the number of training environments increases, as expected. It implies that the value network trained on multiple environments is more likely to memorize the training data and cannot accurately predict the values of unvisited states from the training environments. This memorization problem brings us to train a value network with sufficient regularization.

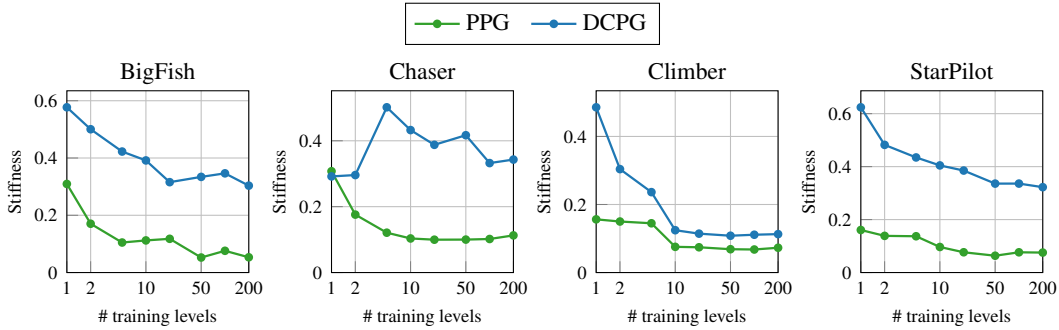
\begin{figure}[t]
    \centering
    \begin{tikzpicture}
        \begin{groupplot}[
		group style = {group size = 4 by 1, horizontal sep=1.1cm},
		width=4.0cm,
		height=4.0cm,
		grid=major,
		xmin=1,
		xmax=200,
		ymin=0,
		scaled x ticks = false,
		tick pos = left,
		tick label style={font=\scriptsize},
		xtick={1, 2, 10, 50, 200},
		xticklabels={1, 2, 10, 50, 200},
		ytick={0, 0.2, 0.4, 0.6, 0.8},
		yticklabels={0, 0.2, 0.4, 0.6, 0.8},
		minor tick style={draw=none},
		xlabel={\# training levels},
		ylabel={Stiffness},
		label style={font=\scriptsize},
		ylabel style={at={(-0.2,0.5)}},
		xlabel style={at={(0.5,-0.15)}},
		title style={font=\small,anchor=north,yshift=8pt},
		legend style={legend columns=5, font=\small, /tikz/every even column/.append style={column sep=0.2cm}},
		legend to name=exp,
        ]
        \pgfplotsinvokeforeach{BigFish,Chaser,Climber,StarPilot}{
        		\nextgroupplot[
			title=#1,
			xmode=log
		]
		\addlegendimage{index of colormap=3 of Paired,mark=*,mark options={scale=0.6},line width=0.30mm}
		\addlegendentry{PPG}
		
		\addlegendimage{index of colormap=1 of Paired,mark=*,mark options={scale=0.6},line width=0.30mm}
		\addlegendentry{DCPG}
		
%

    		\addplot+[index of colormap=3 of Paired,line width=0.30mm,mark=*,mark options={scale=0.6}] table [x=num_train_envs, y=#1_value_grad, col sep=comma]{data/ppg_grad_avg.csv};	
		\addplot+[index of colormap=1 of Paired,line width=0.30mm,mark=*,mark options={scale=0.6}] table [x=num_train_envs, y=#1_value_grad, col sep=comma]{data/appg_grad_avg.csv};	
    		}
        \end{groupplot}
        
	\node[above] at (current bounding box.north) {\pgfplotslegendfromname{exp}};
    \end{tikzpicture}
    \caption{Average stiffness of value networks for PPG and DCPG on 4 Procgen games while varying the number of training levels.}
    \label{fig:cossim_value_selective_both}
\end{figure}

\subsection{Training Value Network with Explicit Regularization}
\label{subsec:regularization}

Next, we examine the effectiveness of value network regularization in the multi-environment setting. We consider applying two existing regularization techniques developed to prevent the memorization problem in the single-environment setting, especially when training data is limited. The first method is discount regularization (DR), which trains a myopic value network with a lower discount factor $\gamma'$ \citep{petrik2008biasing}. The second method is activation regularization (AR), which optimizes a value network with $L_2$ penalty on its outputs:
\begin{align*}
	J_V^\mathrm{reg}(\phi) = \mathbb{E}_{s_t \sim \tau} \left[ \frac{1}{2} \left( V_\phi(s_t) - \hat{R}_t \right)^2  + \frac{\alpha}{2}V_\phi(s_t)^2 \right],
\end{align*}
where $\alpha > 0$ is the regularization coefficient \cite{amit2020discount}. We train PPG agents with each of these two methods using 200 training levels on the Procgen games. We reduce the discount factor from $\gamma = 0.999$ to $\gamma' = 0.995$ for PPG+DR and use $\alpha=0.05$ for PPG+AR. We measure the average training and test returns to evaluate the training performance and its transferability to unseen test environments. 

\begin{figure}[t]
	\centering
    	\begin{subfigure}[b]{0.49\textwidth}
        \centering
        \begin{tikzpicture}
        \begin{groupplot}[
		group style = {group size = 2 by 1, horizontal sep=1.1cm},
    		width=4.0cm,
    		height=4.0cm,
    		grid=major,
    		xmin=0,
    		xmax=2.5e7,
		ymax=33,
    		scaled x ticks = false,
    		tick pos = left,
    		tick label style={font=\scriptsize},
    		xtick={0, 5e6, 10e6, 15e6, 20e6, 25e6},
    		xticklabels={0, 5, 10, 15, 20, 25},
		xlabel={Environment step ($10^6$)},
		ylabel={Average return},
		label style={font=\scriptsize},
		ylabel style={at={(-0.15,0.5)}},
		xlabel style={at={(0.5,-0.15)}},
		legend style={font=\small,legend columns=3, /tikz/every even column/.append style={column sep=0.2cm}},
		legend to name=exp,
        ]
        		\nextgroupplot[
        			title={BigFish, Train},
			title style={font=\small,anchor=north,yshift=10pt},
		]
		\addplot+[index of colormap=3 of Paired,line width=0.2mm] table [x=num_steps, y=train_episode_rewards_mean, col sep=comma]{data/progress-bigfish-ppg.csv};
		\addlegendentry{PPG}
		\addplot+[index of colormap=1 of Paired,line width=0.2mm] table [x=num_steps, y=train_episode_rewards_mean, col sep=comma]{data/progress-bigfish-ppg_discount_0.995.csv};
		\addlegendentry{PPG+DR}
		\addplot+[index of colormap=5 of Paired,line width=0.2mm] table [x=num_steps, y=train_episode_rewards_mean, col sep=comma]{data/progress-bigfish-ppg_norm_0.05.csv};
		\addlegendentry{PPG+AR}
		\coordinate (c1) at (rel axis cs:0.5,1);
		\nextgroupplot[
        			title={BigFish, Test},
			title style={font=\small,anchor=north,yshift=10pt},
		]
		\addplot+[index of colormap=3 of Paired,line width=0.2mm] table [x=num_steps, y=test_episode_rewards_mean, col sep=comma]{data/progress-bigfish-ppg.csv};
		\addlegendentry{PPG}
		\addplot+[index of colormap=1 of Paired,line width=0.2mm] table [x=num_steps, y=test_episode_rewards_mean, col sep=comma]{data/progress-bigfish-ppg_discount_0.995.csv};
		\addlegendentry{PPG+DR}
		\addplot+[index of colormap=5 of Paired,line width=0.2mm] table [x=num_steps, y=test_episode_rewards_mean, col sep=comma]{data/progress-bigfish-ppg_norm_0.05.csv};
		\addlegendentry{PPG+AR}
		 \coordinate (c2) at (rel axis cs:0.5,1);
	\end{groupplot}	
	\coordinate (c3) at ($(c1)!.5!(c2)$);
	\node[above] at (c3 |-current bounding box.north) {\pgfplotslegendfromname{exp}};
        \end{tikzpicture}
        \caption{Average training and test returns.}
        \label{fig:regularization_return}
    	\end{subfigure}
	\hfill
    	\begin{subfigure}[b]{0.49\textwidth}
	\centering
	\begin{tikzpicture}     	
 	\begin{groupplot}[
		group style = {group size = 2 by 1, horizontal sep=1.1cm},
    		width=4.0cm,
    		height=4.0cm,
    		grid=major,
    		xmin=0,
    		xmax=2.5e7,
    		scaled x ticks = false,
    		tick pos = left,
    		tick label style={font=\scriptsize},
    		xtick={0, 5e6, 10e6, 15e6, 20e6, 25e6},
    		xticklabels={0, 5, 10, 15, 20, 25},
		xlabel={Environment step ($10^6$)},
		ylabel={Value},
		label style={font=\scriptsize},
		ylabel style={at={(-0.15,0.5)}},
		xlabel style={at={(0.5,-0.15)}},
		legend style={font=\small,legend columns=2, /tikz/every even column/.append style={column sep=0.2cm}},
		legend to name=exp,
        ]
        \nextgroupplot[
        		title={PPG+DR, BigFish},
		title style={font=\small,anchor=north,yshift=10pt},
	]
	\addplot+[index of colormap=1 of Paired,line width=0.2mm] table [x=num_steps, y=train_ret_n_disc_mean, col sep=comma]{data/progress-bigfish-ppg_discount_0.995.csv};
        \addlegendentry{True value}
        \addplot+[index of colormap=5 of Paired,line width=0.2mm] table [x=num_steps, y=train_init_value_ests_mean, col sep=comma]{data/progress-bigfish-ppg_discount_0.995.csv};
        \addlegendentry{Predicted value}
        	\coordinate (c1) at (rel axis cs:0.5,1);
        \nextgroupplot[
       		title={PPG+AR, BigFish},
		title style={font=\small,anchor=north,yshift=10pt},
	]
	\addplot+[index of colormap=1 of Paired,line width=0.2mm] table [x=num_steps, y=train_ret_n_disc_mean, col sep=comma]{data/progress-bigfish-ppg_norm_0.05.csv};
        \addlegendentry{True}
        \addplot+[index of colormap=5 of Paired,line width=0.2mm] table [x=num_steps, y=train_init_value_ests_mean, col sep=comma]{data/progress-bigfish-ppg_norm_0.05.csv};
        	\coordinate (c2) at (rel axis cs:0.5,1);
        \addlegendentry{Predicted}
        \end{groupplot}	
        \coordinate (c3) at ($(c1)!.5!(c2)$);
        \node[above] at (c3 |-current bounding box.north) {\pgfplotslegendfromname{exp}};
	\end{tikzpicture}
        	\caption{True and predicted values.}
       	\label{fig:regularization_value}
        	\end{subfigure}
        	\caption{(a) Training and test performance curves of PPG, PPG+DR, and PPG+AR on BigFish. (b) True and predicted values measured at the initial states of training environments for PPG+DR and PPG+AR on BigFish. The mean is computed over 10 different runs.}
    \label{fig:regularization_ppg_selective}
\end{figure}
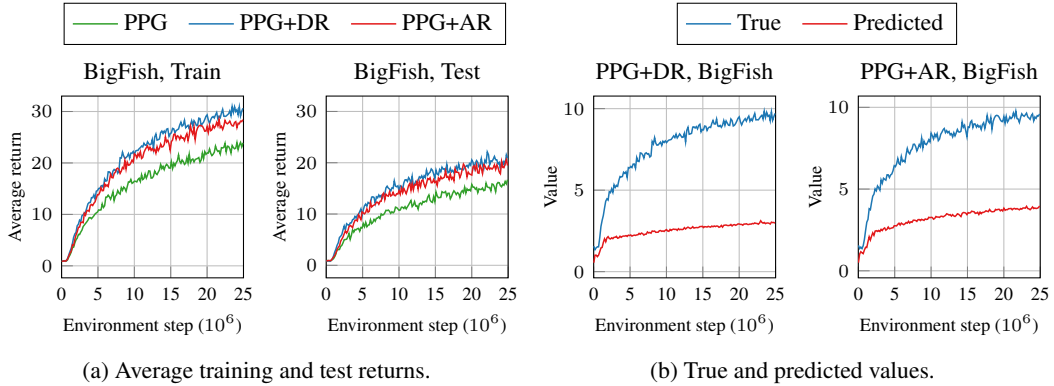

As shown in \Cref{fig:regularization_return}, the value network regularization improves the training and test performance of PPG on BigFish to some extent. It implies that explicitly suppressing the value network also helps to mitigate memorization in the multi-environment setting. We also observe that these regularization methods improve the training and test performance across all Procgen games on average. The detailed experimental setting and results can be found in Appendix B.

Despite its effectiveness, explicit value network regularization can lead to a suboptimal solution as the number of environment steps increases. \Cref{fig:regularization_value} shows the true and predicted values measured at the initial states of the training environments for PPG+DR and PPG+AR on BigFish. The predicted values with explicit regularization reach a plateau too quickly, suggesting that excessive regularization later hinders the value network from learning an accurate value function. This motivates us to develop a more flexible regularization method that boosts training and test performance while allowing the value network to converge to true values.

\section{Delayed-Critic Policy Gradient}
\label{sec:dcpg}

In this section, we present a novel model-free policy gradient algorithm called \emph{Delayed-Critic Policy Gradient} (DCPG), which effectively addresses the memorization problem of the value network in a simple and flexible manner. The key idea is that the value network should be optimized with a larger amount of training data to avoid memorizing a small number of recently visited states, based on the stiffness analysis in \Cref{subsec:memorization}. Furthermore, the value network should be optimized with a delay compared to the policy network to implicitly suppress the value estimate, based on the regularization analysis in \Cref{subsec:regularization}.

\subsection{Algorithm}

DCPG follows a similar procedure to PPG by alternating policy and auxiliary phases. Still, it employs a shared network architecture in the same way as PPO and does not require any additional auxiliary head, as shown in \Cref{fig:network_dcpg}. During the policy phase, which occurs more frequently but with less training data than the auxiliary phase, the policy network is trained with newly-collected trajectories to optimize the policy objective from PPO. In contrast, the value network is constrained to preserve its outputs by optimizing the following value regularizer $C_V$:
\begin{align*}
	C_V(\theta) = \mathbb{E}_{s_t \sim \tau} \left[ \frac{1}{2} \left( V_\theta(s_t) - V_{\theta_\mathrm{old}}(s_t) \right)^2 \right],
\end{align*}
where $V_{\theta_\mathrm{old}}$ is the old value network right before the policy phase. Then, all states and value function targets in the trajectories are stored in a buffer.

During the auxiliary phase, the value network is trained with a larger number of data in the buffer to optimize the value objective from PPO. In contrast, the policy network is constrained to preserve its outputs by optimizing the policy regularizer. Note that since the value network shares the same encoder as the policy network, optimizing the value objective directly plays the role of representation learning for the policy network. Thus, unlike PPG, DCPG does not require an additional auxiliary head for feature distillation. The overall procedure of DCPG can be found in \Cref{alg:dcpg}.

\begin{algorithm}[h]
	\small
	\caption{\cyan{Dynamics-aware} Delayed-Critic Policy Gradient (\cyan{D}DCPG)}
	\begin{algorithmic}[1]
	\Require Policy network $\pi_\theta$, value network $V_\theta$, \cyan{dynamics discriminator $f_\theta$}
    	\For{phase = $1, 2, \ldots $}
    	\State Initialize buffer $\mathcal{B}$
    	\For{iter = $1, 2, \ldots, N_\pi$} \Comment{Policy phase}
    	\State Sample trajectories $\tau$ using $\pi_\theta$ and compute value function target $\hat{R}_t$ for each state $s_t \in \tau$
    	\For{epoch = $1, 2, \ldots, E_\pi$}
    	\State Optimize policy objective $J_\pi(\theta)$ and value regularizer $C_V(\theta)$ with $\tau$
    	\EndFor
	\State Add $(s_t, \cyan{a_t}, \hat{R}_t)$ to $\mathcal{B}$
    	\EndFor
	\For{iter = $1, 2, \ldots, E_\mathrm{aux}$} \Comment{Auxiliary phase}
	\State Optimize value objective $J_V(\theta)$, \cyan{dynamics objective $J_f(\theta)$}, and policy regularizer $C_\pi(\theta)$ with $\mathcal{B}$
	\EndFor
    	\EndFor
	\end{algorithmic}
	\label{alg:dcpg}
\end{algorithm}

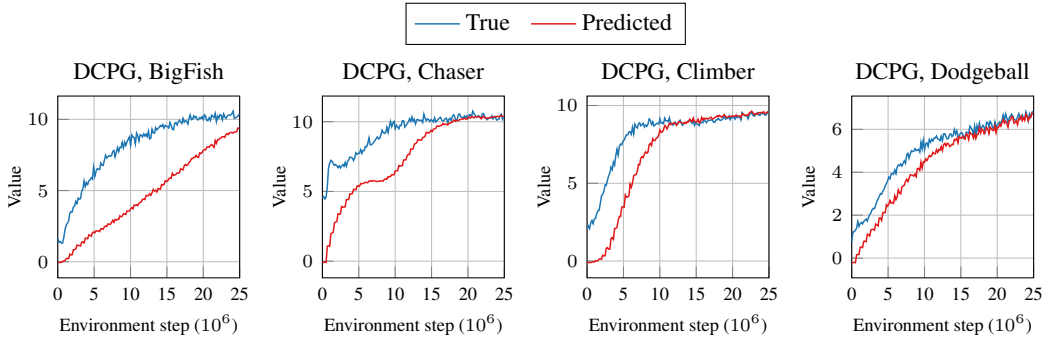
\begin{figure}[t]
    \centering
    \begin{tikzpicture}     	
 	\begin{groupplot}[
		group style = {group size = 4 by 1, horizontal sep=1.1cm},
    		width=4.0cm,
    		height=4.0cm,
    		grid=major,
    		xmin=0,
    		xmax=2.5e7,
    		scaled x ticks = false,
    		tick pos = left,
    		tick label style={font=\scriptsize},
    		xtick={0, 5e6, 10e6, 15e6, 20e6, 25e6},
    		xticklabels={0, 5, 10, 15, 20, 25},
		xlabel={Environment step ($10^6$)},
		ylabel={Value},
		label style={font=\scriptsize},
		ylabel style={at={(-0.15,0.5)}},
		xlabel style={at={(0.5,-0.15)}},
		legend style={font=\small,legend columns=2, /tikz/every even column/.append style={column sep=0.2cm}},
		legend to name=exp,
        ]
            \nextgroupplot[
            		title={DCPG, BigFish},
    		title style={font=\small,anchor=north,yshift=10pt},
    	]
    	 \addplot+[index of colormap=1 of Paired,line width=0.2mm] table [x=num_steps, y=train_ret_n_disc_mean, col sep=comma]{data/progress-bigfish-appg.csv};
            \addlegendentry{True value}
            \addplot+[index of colormap=5 of Paired,line width=0.2mm] table [x=num_steps, y=train_init_value_ests_mean, col sep=comma]{data/progress-bigfish-appg.csv};
            \addlegendentry{Predicted value}
            	\coordinate (c1) at (rel axis cs:0.5,1);
            \nextgroupplot[
           		title={DCPG, Chaser},
    		title style={font=\small,anchor=north,yshift=10pt},
    	]
    	 \addplot+[index of colormap=1 of Paired,line width=0.2mm] table [x=num_steps, y=train_ret_n_disc_mean, col sep=comma]{data/progress-chaser-appg.csv};
            \addlegendentry{True value}
            \addplot+[index of colormap=5 of Paired,line width=0.2mm] table [x=num_steps, y=train_init_value_ests_mean, col sep=comma]{data/progress-chaser-appg.csv};
            \addlegendentry{Predicted value}
            \nextgroupplot[
           	title={DCPG, Climber},
    		title style={font=\small,anchor=north,yshift=10pt},
    	]
    	 \addplot+[index of colormap=1 of Paired,line width=0.2mm] table [x=num_steps, y=train_ret_n_disc_mean, col sep=comma]{data/progress-climber-appg.csv};
            \addlegendentry{True value}
            \addplot+[index of colormap=5 of Paired,line width=0.2mm] table [x=num_steps, y=train_init_value_ests_mean, col sep=comma]{data/progress-climber-appg.csv};
            \addlegendentry{Predicted value}
           \nextgroupplot[
           		title={DCPG, Dodgeball},
    		title style={font=\small,anchor=north,yshift=10pt},
    	]
    	 \addplot+[index of colormap=1 of Paired,line width=0.2mm] table [x=num_steps, y=train_ret_n_disc_mean, col sep=comma]{data/progress-dodgeball-appg.csv};
            \addlegendentry{True}
            \addplot+[index of colormap=5 of Paired,line width=0.2mm] table [x=num_steps, y=train_init_value_ests_mean, col sep=comma]{data/progress-dodgeball-appg.csv};
            	\coordinate (c2) at (rel axis cs:0.5,1);
            \addlegendentry{Predicted}
        \end{groupplot}	
        \coordinate (c3) at ($(c1)!.5!(c2)$);
        \node[above] at (c3 |-current bounding box.north) {\pgfplotslegendfromname{exp}};
	\end{tikzpicture}
        	\caption{True and predicted values measured at the initial states of training environments for DCPG agents on 4 Procgen games. The mean is computed over 10 different runs.}
       	\label{fig:regularization_dcpg_selective}
\end{figure}

\subsection{Value Network Analysis of Delayed Critic Update}

The delayed critic update in DCPG can function as implicit regularization of the value network. Since the policy improvement is not immediately reflected in the value network, it encourages smaller value estimates than the true values. To validate this claim, we train DCPG agents using 200 training levels on the Procgen games and measure the true and predicted values at the initial states of the training levels. \Cref{fig:regularization_dcpg_selective} shows the value estimates are kept lower than the true values at the early stage of training but recover the true values as training progresses. More results can be found in Appendix C.

We also evaluate the effectiveness of the delayed critic update in mitigating the memorization problem by measuring the stiffness of the value network as done in \Cref{subsec:memorization}. The blue lines in \Cref{fig:cossim_value_selective_both} show that DCPG has higher stiffness than PPG, implying that the value network of DCPG is more robust against memorization. We conclude that the delayed critic update serves as an effective regularizer for the value network. More results can be found in Appendix A.

\subsection{Learning Forward and Inverse Dynamics with Single Discriminator}

In the context of multi-task learning, learning an auxiliary task can operate as a good regularizer that prevents memorization in the main task \cite{ruder2017overview}. Motivated by this, we consider learning an auxiliary task using the representations from the encoder and jointly learning the value network with the auxiliary objective. Suppose the buffer $\mathcal{B}$ used in the auxiliary phase contains a transition tuple $(s_t, a_t, s_{t+1})$. Since the transition function from each environment is assumed to have the same underlying structure, it is natural to train a neural network that models the forward dynamics of environments in the latent space of the encoder. Concretely, a dynamics head $f_\theta$ takes the embedding of the current state $s_t$ and the current action $a_t$ as inputs and predicts the embedding of the next state $s_t$. However, it is well known that a forward dynamics model using a neural network tends to overfit a small number of training data and make poor predictions on unseen states \cite{chua2018deep}. To address this, we consider learning the forward dynamics by training a discriminator that determines whether the next state is valid, given the current state and action:
\begin{align*}
	f_\theta(s_t, a_t, s_{t+1}) = 1, ~ f_\theta(s_t, a_t, \hat{s}_{t+1}) = 0, \ \hat{s}_{t+1} \in \mathcal{B} \setminus \{s_{t+1}\}.
\end{align*}
Such a discriminator might discard information about the action to distinguish whether a transition is valid. For example, the discriminator can predict the valid transition by using only the proximity of the current and next states (see Appendix F for more details). To avoid this, we train the discriminator to jointly distinguish whether the current action is valid, given the current and next states (\ie, inverse dynamics):
\begin{align*}
	f_\theta(s_t, a_t, s_{t+1}) = 1, ~ f_\theta(s_t, \hat{a}_t, s_{t+1}) = 0, \ \hat{a}_t \in \mathcal{B} \setminus \{a_t\}.	
\end{align*}
Finally, we optimize the following dynamics objective during the auxiliary phase:
\begin{align*}
	J_f(\theta) = \mathbb{E}_{s_t, a_t, s_{t+1} \sim \mathcal{B}} \big[ \log \left(f_\theta(s_t, a_t, {s}_{t+1})\right) &+ \log \left(1 - f_\theta(s_t, a_t, \hat{s}_{t+1}) \right)  \\
	&+\eta \log \left( 1 - f_\theta(s_t, \hat{a}_t, s_{t+1}) \right) \big],
\end{align*}
where $\hat{s}_{t+1} \sim \mathcal{U}(\mathcal{B} \setminus \{ s_{t+1} \})$, $\hat{a}_t \sim \mathcal{U}(\mathcal{B} \setminus \{ a_t \})$, and $\eta > 0$ is the coefficient for the inverse dynamics. We name the resulting algorithm \emph{Dynamics-aware Delayed-Critic Policy Gradient} (DDCPG). The network architecture and algorithm for DDCPG are shown in \Cref{fig:network_dcpg} and \Cref{alg:dcpg}, respectively (the differences with DCPG are marked in cyan).

\section{Experiments}
\label{sec:exp}

\subsection{Experimental Settings}

We evaluate the observational generalization performance of our methods on the Procgen benchmark. We use the ``easy'' difficulty mode, which most prior works have focused on. We train agents on 200 training levels generated by seeds from 0 to 200 for 25M environment steps and test them on held-out test levels, following the practice in \citet{cobbe2020leveraging}. We measure the average return of 100 test episodes and report its mean and standard deviation over 10 runs with different initialization.

We compare our methods with PPO and 4 other baselines that use PPO as the backbone algorithm: UCB-DrAC, PLR, PPG, and DAAC \citep{raileanu2021automatic,jiang2021prioritized,cobbe2021phasic,raileanu2021decoupling}. UCB-DrAC is a data augmentation algorithm for training policy and value networks to be robust against various transformations. PLR is a sampling algorithm for the procedural generation that selects the next training level based on its future learning potential. PPG is the previous state-of-the-art method on the Procgen benchmark. DAAC is motivated by PPG and distills the advantage function into the policy network instead. For each method, we use the same set of hyperparameters across all games, following the standard protocol in the ALE and Procgen benchmarks \citep{mnih2015human,cobbe2020leveraging}. For our methods, we use the same hyperparameter setting as PPG for a fair comparison. To compare the performance of each method with a single score, we calculate the PPO-normalized score averaged over all Procgen games, which is computed by dividing the average return of each method by PPO, following the practice in \citet{jiang2021prioritized,raileanu2021automatic}. For the implementation details and hyperparameters, please refer to Appendix D. The code can be found at \url{https://github.com/snu-mllab/DCPG}.

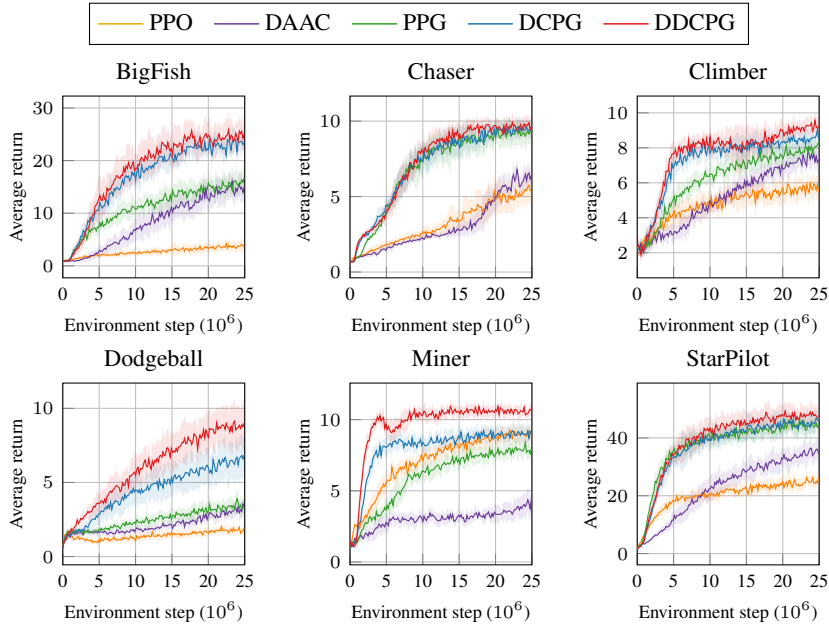
\begin{figure}[t]
    \centering
    \begin{tikzpicture}
        \begin{groupplot}[
		group style = {group name=procgen, group size = 3 by 2, horizontal sep=1.4cm, vertical sep=1.4cm},
		width=4cm,
		height=4cm,
		grid=major,
		xmin=0,
		xmax=2.5e7,
		scaled x ticks = false,
		tick pos = left,
		tick label style={font=\scriptsize},
		xtick={0, 5e6, 10e6, 15e6, 20e6, 25e6},
		xticklabels={0, 5, 10, 15, 20, 25},
		xlabel={Environment step ($10^6$)},
		ylabel={Average return},
		label style={font=\scriptsize},
		ylabel style={at={(-0.15,0.5)}},
		xlabel style={at={(0.5,-0.15)}},
		title style={font=\small,anchor=north,yshift=10pt},
		legend style={font=\small,legend columns=7, /tikz/every even column/.append style={column sep=0.2cm}},
		legend to name=exp,
        ]
		\pgfplotsinvokeforeach{BigFish,Chaser,Climber,Dodgeball,Miner,StarPilot}{
		    \nextgroupplot[
		        title=#1,        
		    ]
		\addlegendimage{index of colormap=5 of Dark2,line width=0.2mm}
		\addlegendentry{PPO}
		\addlegendimage{index of colormap=9 of Paired,line width=0.2mm}
    		\addlegendentry{DAAC}
		\addlegendimage{index of colormap=3 of Paired,line width=0.2mm}
    		\addlegendentry{PPG}
		\addlegendimage{index of colormap=1 of Paired,line width=0.2mm}
    		\addlegendentry{DCPG}
		\addlegendimage{index of colormap=5 of Paired,line width=0.2mm}
    		\addlegendentry{DDCPG}
    		\addplot+[index of colormap=7 of Paired,line width=0.15mm] table [x=num_steps, y=test_episode_rewards_mean, col sep=comma]{data/progress-#1-ppo.csv};
    		\addplot+[index of colormap=9 of Paired,line width=0.15mm] table [x=num_steps, y=test_episode_rewards_mean, col sep=comma]{data/progress-#1-daac.csv};
    		\addplot+[index of colormap=3 of Paired,line width=0.15mm] table [x=num_steps, y=test_episode_rewards_mean, col sep=comma]{data/progress-#1-ppg.csv};
    		\addplot+[index of colormap=1 of Paired,line width=0.15mm] table [x=num_steps, y=test_episode_rewards_mean, col sep=comma]{data/progress-#1-appg.csv};
    		\addplot+[index of colormap=5 of Paired,line width=0.15mm] table [x=num_steps, y=test_episode_rewards_mean, col sep=comma]{data/progress-#1-ssl_v1.csv};
    		\addplot[name path=plus,draw=none] table [x=num_steps, y=test_episode_rewards_mean_plus_std, col sep=comma]{data/progress-#1-ppo.csv};
    		\addplot[name path=minus,draw=none] table [x=num_steps, y=test_episode_rewards_mean_minus_std, col sep=comma]{data/progress-#1-ppo.csv};
    		\addplot[index of colormap=7 of Paired,fill opacity=0.12] fill between[of=plus and minus];
    		\addplot[name path=plus,draw=none] table [x=num_steps, y=test_episode_rewards_mean_plus_std, col sep=comma]{data/progress-#1-daac.csv};
    		\addplot[name path=minus,draw=none] table [x=num_steps, y=test_episode_rewards_mean_minus_std, col sep=comma]{data/progress-#1-daac.csv};
    		\addplot[index of colormap=9 of Paired,fill opacity=0.12] fill between[of=plus and minus];
    		\addplot[name path=plus,draw=none] table [x=num_steps, y=test_episode_rewards_mean_plus_std, col sep=comma]{data/progress-#1-ppg.csv};
    		\addplot[name path=minus,draw=none] table [x=num_steps, y=test_episode_rewards_mean_minus_std, col sep=comma]{data/progress-#1-ppg.csv};
    		\addplot[index of colormap=3 of Paired,fill opacity=0.12] fill between[of=plus and minus];
    		\addplot[name path=plus,draw=none] table [x=num_steps, y=test_episode_rewards_mean_plus_std, col sep=comma]{data/progress-#1-appg.csv};
    		\addplot[name path=minus,draw=none] table [x=num_steps, y=test_episode_rewards_mean_minus_std, col sep=comma]{data/progress-#1-appg.csv};
    		\addplot[index of colormap=1 of Paired,fill opacity=0.12] fill between[of=plus and minus];
    		\addplot[name path=plus,draw=none] table [x=num_steps, y=test_episode_rewards_mean_plus_std, col sep=comma]{data/progress-#1-ssl_v1.csv};
    		\addplot[name path=minus,draw=none] table [x=num_steps, y=test_episode_rewards_mean_minus_std, col sep=comma]{data/progress-#1-ssl_v1.csv};
    		\addplot[index of colormap=5 of Paired,fill opacity=0.12] fill between[of=plus and minus];
		}
        \end{groupplot}
	\node[above] at (current bounding box.north) {\pgfplotslegendfromname{exp}};
    \end{tikzpicture}
    \caption{Test performance curves of each method on 6 Procgen games. Each agent is trained on 200 training levels for 25M environment steps and evaluated on 100 unseen test levels. The mean and standard deviation are computed over 10 different runs.}
    \label{fig:procgen_test_selective}
\end{figure}

\subsection{Observational Generalization Performance on Procgen Benchmark}

\Cref{tab:procgen_test} shows the average test returns of each method on all 16 Procgen games. DCPG significantly outperforms all the baselines, yielding a 24\%p improvement on average compared to the previous state-of-the-art PPG. Adding dynamics learning to DCPG can further improve the test performance with an 18\%p increase in the PPO-normalized score. We also provide the average training returns on all Procgen games in Appendix E, showing that our methods achieve better sample efficiency in the multi-environment setting. Furthermore, we evaluate the performance of our methods using the Min-Max normalized score and report mean, median, and interquartile mean (IQM) scores in Appendix G, following the recent practice in \citet{agarwal2021deep}. We observe that the performance improvements of our methods are statistically significant for all the metrics we consider.

\begin{table}[t]
    \centering
    \caption{Average test returns of each method on all 16 Procgen games. Each agent is trained on 200 training levels for 25M environment steps and evaluated on 100 unseen test levels. The mean and standard deviation are computed over 10 different runs.}
    \adjustbox{max width=\textwidth}{
    \begin{tabular}{l|rrrrr|rr}
        \toprule
        Environment & PPO & UCB-DrAC & PLR & DAAC\tablefootnote{Results reported in the original paper use a different hyperparameter setting for each game.} & PPG & DCPG & DDCPG \\
        \midrule
        BigFish & 3.4 $\pm$ 1.0 & 6.6 $\pm$ 2.5 & 10.8 $\pm$ 2.5 & 15.3 $\pm$ 3.0 & 16.4 $\pm$ 2.0 & 23.1 $\pm$ 2.3 & 25.9 $\pm$ 2.3 \\
        BossFight & 7.4 $\pm$ 0.7 & 7.2 $\pm$ 1.0 & 8.8 $\pm$ 0.7 & 9.5 $\pm$ 0.8 & 10.4 $\pm$ 0.6 & 10.2 $\pm$ 0.4 & 10.6 $\pm$ 0.5 \\
        CaveFlyer & 5.2 $\pm$ 0.6 & 4.4 $\pm$ 0.8 & 6.3 $\pm$ 0.7 & 5.1 $\pm$ 0.6 & 7.7 $\pm$ 0.7 & 6.3 $\pm$ 0.4 & 6.0 $\pm$ 0.3 \\
        Chaser & 5.5 $\pm$ 0.8 & 6.7 $\pm$ 0.5 & 7.5 $\pm$ 0.8 & 6.3 $\pm$ 0.8 & 8.9 $\pm$ 0.9 & 9.7 $\pm$ 0.4 & 9.9 $\pm$ 0.6 \\
        Climber & 5.5 $\pm$ 0.6 & 6.2 $\pm$ 0.6 & 6.4 $\pm$ 0.5 & 7.4 $\pm$ 0.5 & 8.1 $\pm$ 0.4 & 8.5 $\pm$ 0.7 & 9.3 $\pm$ 0.7  \\
        CoinRun & 8.8 $\pm$ 0.3 & 8.6 $\pm$ 0.4 & 8.9 $\pm$ 0.3 & 9.3 $\pm$ 0.2 & 8.9 $\pm$ 0.3 & 8.5 $\pm$ 0.4 & 8.4 $\pm$ 0.3 \\
        Dodgeball & 1.9 $\pm$ 0.3 & 4.5 $\pm$ 1.4 & 2.2 $\pm$ 0.4 & 3.2 $\pm$ 0.5 & 3.6 $\pm$ 0.7 & 6.7 $\pm$ 0.4 & 9.0 $\pm$ 1.6 \\
        FruitBot & 27.5 $\pm$ 1.5 & 26.9 $\pm$ 1.7 & 27.7 $\pm$ 1.3 & 28.5 $\pm$ 1.4 & 29.1 $\pm$ 1.0 & 29.0 $\pm$ 1.2 & 28.8 $\pm$ 0.6 \\
        Heist & 2.6 $\pm$ 0.6 & 3.4 $\pm$ 1.0 & 3.1 $\pm$ 0.6 & 3.0 $\pm$ 0.5 & 2.8 $\pm$ 0.5 & 3.3 $\pm$ 0.5 & 4.4 $\pm$ 0.8 \\
        Jumper & 5.7 $\pm$ 0.4 & 5.6 $\pm$ 0.4 & 5.9 $\pm$ 0.4 & 5.8 $\pm$ 0.4 & 6.0 $\pm$ 0.4 & 6.2 $\pm$ 0.5 & 6.4 $\pm$ 0.4 \\
        Leaper & 5.6 $\pm$ 1.5 & 3.6 $\pm$ 0.7 & 7.3 $\pm$ 0.2 & 8.2 $\pm$ 1.1 & 7.2 $\pm$ 2.4 & 6.9 $\pm$ 1.7 & 6.8 $\pm$ 1.5 \\
        Maze & 5.3 $\pm$ 0.7 & 5.7 $\pm$ 1.1 & 5.6 $\pm$ 0.5 & 5.6 $\pm$ 0.6 & 5.3 $\pm$ 0.4 & 5.7 $\pm$ 0.7 & 6.5 $\pm$ 0.6 \\
        Miner & 9.2 $\pm$ 0.6 & 10.1 $\pm$ 0.7 & 9.4 $\pm$ 0.7 & 4.4 $\pm$ 1.1 & 7.6 $\pm$ 0.7 & 9.1 $\pm$ 0.8 & 10.6 $\pm$ 0.5 \\
        Ninja & 5.7 $\pm$ 0.5 & 5.7 $\pm$ 0.5 & 7.1 $\pm$ 0.5 & 6.7 $\pm$ 0.7 & 6.7 $\pm$ 0.4 & 6.3 $\pm$ 0.6 & 6.6 $\pm$ 0.5 \\
        Plunder & 5.0 $\pm$ 0.5 & 8.1 $\pm$ 1.5 & 8.7 $\pm$ 1.3 & 5.6 $\pm$ 0.6 & 13.7 $\pm$ 1.7 & 13.7 $\pm$ 1.3 & 12.9 $\pm$ 2.5 \\
        StarPilot & 25.8 $\pm$ 2.0 & 27.0 $\pm$ 3.2 & 24.9 $\pm$ 3.6 & 35.9 $\pm$ 3.3 & 44.8 $\pm$ 2.6 & 46.1 $\pm$ 2.1 & 46.6 $\pm$ 4.1 \\
        \midrule
        PPO-norm score (\%) & 100.0 $\pm$ 3.1 & 120.7 $\pm$ 10.7 & 130.0 $\pm$ 6.0 & 136.7 $\pm$ 7.6 & 160.3 $\pm$ 6.3 & \textbf{184.5} $\pm$ \textbf{5.2} & \textbf{202.2} $\pm$ \textbf{10.2} \\
        \bottomrule
    \end{tabular}
    }
    \label{tab:procgen_test}
\end{table}

\Cref{fig:procgen_test_selective} shows the test performance curves of each method on 6 Procgen games throughout training. Our methods achieve superior final test performance and acquire generalization ability with fewer environment interactions. For example, our methods reach the final performance of PPG using only 20\% of total environment steps on BigFish and Climber and 10\% on Dodgeball and Miner. More experimental results can be found in Appendix E.

\subsection{Ablation Study for Delayed Critic Update}

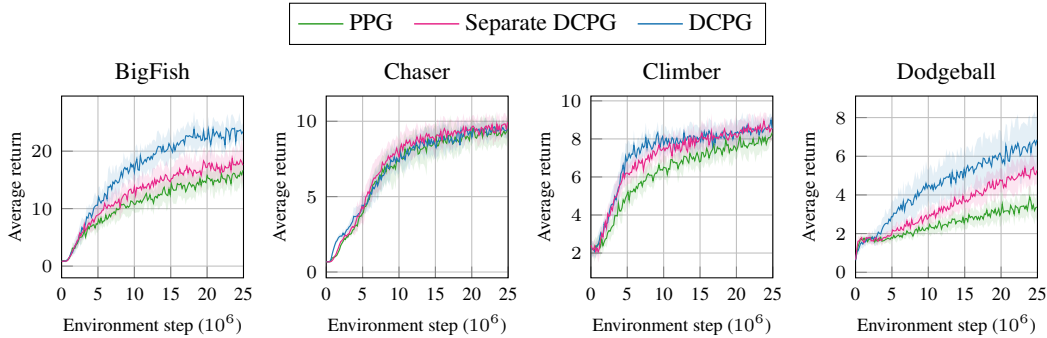
\begin{figure}[t]
    \centering
    \begin{tikzpicture}
        \begin{groupplot}[
		group style = {group name=procgen, group size = 4 by 1, horizontal sep=1.1cm},
		width=4cm,
		height=4cm,
		grid=major,
		xmin=0,
		xmax=2.5e7,
		scaled x ticks = false,
		tick pos = left,
		tick label style={font=\scriptsize},
		xtick={0, 5e6, 10e6, 15e6, 20e6, 25e6},
		xticklabels={0, 5, 10, 15, 20, 25},
		xlabel={Environment step ($10^6$)},
		ylabel={Average return},
		label style={font=\scriptsize},
		ylabel style={at={(-0.15,0.5)}},
		xlabel style={at={(0.5,-0.15)}},
		title style={font=\small,anchor=north,yshift=10pt},
		legend style={font=\small,legend columns=7, /tikz/every even column/.append style={column sep=0.2cm}},
		legend to name=exp,
        ]
		\pgfplotsinvokeforeach{BigFish,Chaser,Climber,Dodgeball}{
		    \nextgroupplot[
		        title=#1,        
		    ]
		\addlegendimage{index of colormap=3 of Paired,line width=0.2mm}
    		\addlegendentry{PPG}
		\addlegendimage{index of colormap=3 of Dark2,line width=0.2mm}
    		\addlegendentry{Separate DCPG}
		\addlegendimage{index of colormap=1 of Paired,line width=0.2mm}
    		\addlegendentry{DCPG}	
    		\addplot+[index of colormap=3 of Paired,line width=0.15mm] table [x=num_steps, y=test_episode_rewards_mean, col sep=comma]{data/progress-#1-ppg.csv};
    		\addplot+[index of colormap=1 of Paired,line width=0.15mm] table [x=num_steps, y=test_episode_rewards_mean, col sep=comma]{data/progress-#1-appg.csv};
    		\addplot+[index of colormap=3 of Dark2,line width=0.15mm] table [x=num_steps, y=test_episode_rewards_mean, col sep=comma]{data/progress-#1-appg_sep.csv};
    		\addplot[name path=plus,draw=none] table [x=num_steps, y=test_episode_rewards_mean_plus_std, col sep=comma]{data/progress-#1-ppg.csv};
    		\addplot[name path=minus,draw=none] table [x=num_steps, y=test_episode_rewards_mean_minus_std, col sep=comma]{data/progress-#1-ppg.csv};
    		\addplot[index of colormap=3 of Paired,fill opacity=0.12] fill between[of=plus and minus];
    		\addplot[name path=plus,draw=none] table [x=num_steps, y=test_episode_rewards_mean_plus_std, col sep=comma]{data/progress-#1-appg.csv};
    		\addplot[name path=minus,draw=none] table [x=num_steps, y=test_episode_rewards_mean_minus_std, col sep=comma]{data/progress-#1-appg.csv};
    		\addplot[index of colormap=1 of Paired,fill opacity=0.12] fill between[of=plus and minus];
    		\addplot[name path=plus,draw=none] table [x=num_steps, y=test_episode_rewards_mean_plus_std, col sep=comma]{data/progress-#1-appg_sep.csv};
    		\addplot[name path=minus,draw=none] table [x=num_steps, y=test_episode_rewards_mean_minus_std, col sep=comma]{data/progress-#1-appg_sep.csv};
    		\addplot[index of colormap=3 of Dark2,fill opacity=0.12] fill between[of=plus and minus];
    		}
        \end{groupplot}
	\node[above] at (current bounding box.north) {\pgfplotslegendfromname{exp}};
    \end{tikzpicture}
    \caption{Test performance curves of PPG, Separate DCPG, and DCPG on 4 Procgen games. Each agent is trained on 200 training levels for 25M environment steps and evaluated on 100 unseen test levels. The mean and standard deviation are computed over 10 different runs.}
    \label{fig:procgen_test_ablation}
\end{figure}

The delayed critic update not only alleviates the memorization problem but also provides better state representations for generalization to the policy network. To validate this, we train DCPG agents with an additional value network $V_\phi$ using a separate encoder (Separate DCPG). The separate value network $V_\phi$ is aggressively optimized in the same way as PPG and only used for calculating advantages in the policy objective. The original value network $V_\theta$ is only used for learning representations for the policy network with delayed updates. \Cref{fig:procgen_test_ablation} shows that Separate DCPG achieves better test performance than PPG, implying that the delayed critic update produces more generalizable representations. For the detailed experimental settings and results, please refer to Appendix F.

\subsection{Ablation Study for Dynamics Learning}

We conduct an ablation study to show the effectiveness of training forward and inverse dynamics using a single discriminator. We train DCPG agents with only forward dynamics learning (DCPG+F) and only inverse dynamics learning (DCPG+I). In addition, we train DCPG agents with forward and inverse dynamics learning using two separate discriminators (DCPG+FI). \Cref{tab:procgen_test_ssl} shows that DDCPG, which jointly optimizes the forward and inverse dynamics with a single discriminator, achieves better test performance than the other methods with dynamics learning. Note that inverse dynamics provides performance gain only when jointly trained with forward dynamics using a single discriminator. We provide the detailed experimental settings and results in Appendix F.

\begin{table}[h]
	\small
	\centering
	\caption{PPO-normalized score of each method with dynamics learning on all 16 Procgen games. Each agent is trained on 200 training levels for 25M environment steps and evaluated on 100 unseen test levels. The mean and standard deviation are computed over 10 different runs.}
    	\begin{tabular}{l|rrrrr}
	\toprule
	 & DCPG & DCPG+F & DCPG+I & DCPG+FI & DDCPG \\
	\midrule
	PPO-norm score (\%) & 184.5 $\pm$ 5.2 & 195.9 $\pm$ 7.7 & 185.5 $\pm$ 6.5 & 194.6 $\pm$ 6.2 & \textbf{202.2} $\pm$ \textbf{10.2} \\
	\bottomrule
	\end{tabular}
    	\label{tab:procgen_test_ssl}
\end{table}

\section{Related Works}

\paragraph{Observational generalization in RL} One prominent approach for improving observational generalization in model-free RL is to employ regularization techniques developed for supervised learning, such as batch normalization, weight regularization, and information bottleneck \citep{farebrother2018generalization,cobbe2019quantifying,igl2019generalization}. Several works adopt data augmentation strategies commonly used in computer vision to address the generalization problem \citep{laskin2020reinforcement,yarats2021image}. For example, UCB-DrAC proposes a method that automatically finds an effective augmentation and introduces new regularization terms to learn robust state representations under various image transformations \citep{raileanu2021automatic}. Recently, it has been shown that MuZero Reanalyse, the state-of-the-art model-based RL algorithm, achieves better observational generalization performance than model-free approaches, though it requires a larger network architecture \citep{anand2022procedural}.

Another line of work uses self-supervised learning to learn invariant representations that can generalize to unseen environments. Some methods attempt to learn state representations that capture long-term behavior proximity between states using bisimulation or policy similarity metrics to discard irrelevant information for generalization \cite{zhang2021learning,agarwal2021contrastive,mazoure2022crosstrajectory}. The idea of learning dynamics as a self-supervised task, originally developed for sample efficiency, has also been used to improve observational generalization \citep{gelada2019deepmdp,schwarzer2021dataefficient}. DIM learns a state representation that can predict the representations of successive timesteps using self-supervised learning \citep{mazoure2020deep}. It has also been tested that predicting future state representations conditioned on the current state and action improves the generalization performance of MuZero \cite{anand2022procedural}. Our work aims to enhance dynamics learning further by jointly modeling the forward and inverse dynamics objectives using a single discriminator.

PPG is most closely related to our work, which proposes a better training algorithm to improve both observational generalization and sample efficiency in the multi-environment setting \citep{cobbe2021phasic}. It trains the policy and value networks independently using a separate architecture and different levels of sample reuse to avoid interference and obtain a better-trained value network. DAAC also uses decoupled policy and value networks for a more accurate value function but distills the advantage function into the policy encoder under the assumption that it is less prone to overfitting to environment-specific features than the value function  \citep{raileanu2021decoupling}. In contrast, we focus on regularizing the value network while obtaining better representation, which distinguishes our work from the prior methods.

\paragraph{Learning value networks with regularization}
Discount regularization is one common regularization method to mitigate overfitting in the value network \citep{BertsekasTsitsiklis96}. A lower discount factor can lead to better performance in approximated dynamic programming when the approximation error is large \citep{petrik2008biasing}. It is also helpful for preventing overfitting when dealing with limited data in POMDPs \citep{franccois2019overfitting}. Activation regularization penalizes the outputs of the value network and has a regularization effect similar to that of discount regularization \citep{amit2020discount}. It has been shown that activation regularization is equivalent to discount regularization in batch TD learning and also effective in online RL \citep{amit2020discount}.

\section{Conclusion}

We have investigated the difficulty of learning an accurate value network in the multi-environment setting and shown that suppressing the value network with appropriate regularization can be helpful to improve both the training and test performance. Based on this observation, we propose Delayed-Critic Policy Gradient (DCPG), where the value network is implicitly regularized to have lower predicted values at the early stage of training with delayed updates compared to the policy network. We find that the delayed value update prevents the memorization of training data and produces more generalizable representations that can be extended to unseen test environments. Furthermore, we introduce a simple self-supervised task that jointly learns forward and inverse dynamics using a single discriminator, which can be easily combined with DCPG. Our algorithms exhibit state-of-the-art performance in observational generalization on the challenging Procgen benchmark.

\section*{Acknowledgements}

This work was supported in part by Samsung Advanced Institute of Technology, Samsung Electronics
Co., Ltd., Institute of Information \& Communications Technology Planning \& Evaluation (IITP) grant
funded by the Korea government (MSIT) (No. 2019-0-01371, Development of brain-inspired AI with human-like intelligence, No. 2020-0-00882, (SW STAR LAB) Development of deployable learning intelligence via self-sustainable and trustworthy machine learning, and No. 2022-0-00480, Development of Training and Inference Methods for Goal-Oriented Artificial Intelligence Agents), and a grant of the Korea Health Technology R\&D Project through the Korea Health Industry Development Institute (KHIDI), funded by the Ministry of Health \& Welfare, Republic of Korea (grant number: HI21C1074). This material is based upon work supported by the Air Force Office of Scientific Research under award number FA2386-22-1-4010. Hyun Oh Song is the corresponding author.

\bibliography{main}
\bibliographystyle{plainnat}

\section*{Checklist}

\begin{enumerate}

\item For all authors...
\begin{enumerate}
  \item Do the main claims made in the abstract and introduction accurately reflect the paper's contributions and scope?
    \answerYes{We clearly reflect the main claims in \Cref{sec:motivation} and \Cref{sec:dcpg}.}
  \item Did you describe the limitations of your work?
    \answerNA{}
  \item Did you discuss any potential negative societal impacts of your work?
    \answerNA{}
  \item Have you read the ethics review guidelines and ensured that your paper conforms to them?
    \answerYes{We have read the ethics review guidelines and checked our paper has no conflict.}
\end{enumerate}

\item If you are including theoretical results...
\begin{enumerate}
  \item Did you state the full set of assumptions of all theoretical results?
    \answerNA{}
        \item Did you include complete proofs of all theoretical results?
    \answerNA{}
\end{enumerate}

\item If you ran experiments...
\begin{enumerate}
  \item Did you include the code, data, and instructions needed to reproduce the main experimental results (either in the supplemental material or as a URL)?
    \answerYes{We provide the code for reproducing the main results in the supplementary material.}
  \item Did you specify all the training details (e.g., data splits, hyperparameters, how they were chosen)?
    \answerYes{We provide the implementation details and hyperparameters of all baselines and our methods in the supplementary material.}
        \item Did you report error bars (e.g., with respect to the random seed after running experiments multiple times)?
    \answerYes{We conduct experiments over 10 training seeds and report mean and standard deviation.}
        \item Did you include the total amount of compute and the type of resources used (e.g., type of GPUs, internal cluster, or cloud provider)?
    \answerYes{We describe the computational resource in the supplementary material.}
\end{enumerate}

\item If you are using existing assets (e.g., code, data, models) or curating/releasing new assets...
\begin{enumerate}
  \item If your work uses existing assets, did you cite the creators?
    \answerYes{We cite the benchmark used for our experiments.}
  \item Did you mention the license of the assets?
    \answerNA{}
  \item Did you include any new assets either in the supplemental material or as a URL?
    \answerNA{}
  \item Did you discuss whether and how consent was obtained from people whose data you're using/curating?
    \answerNA{}
  \item Did you discuss whether the data you are using/curating contains personally identifiable information or offensive content?
    \answerNA{}
\end{enumerate}

\item If you used crowdsourcing or conducted research with human subjects...
\begin{enumerate}
  \item Did you include the full text of instructions given to participants and screenshots, if applicable?
    \answerNA{}
  \item Did you describe any potential participant risks, with links to Institutional Review Board (IRB) approvals, if applicable?
    \answerNA{}
  \item Did you include the estimated hourly wage paid to participants and the total amount spent on participant compensation?
    \answerNA{}
\end{enumerate}

\end{enumerate}

\newpage

\appendix

\section{Stiffness Analysis}

To quantify how much the number of training environments affects the memorization of the value network, we train PPG agents using the training levels of seeds from 0 to $n$ for 8M environment steps on all 16 Procgen games and measure the stiffness of the value network between states while varying the number of training environments $n \in \{1, 2, 5, 10, 20, 50, 100, 200\}$. Throughout the training, we sample trajectories from the training levels, which consist of $2^{14}$ (=16,384) states, and compute the individual gradient of the value objective for each state using BackPACK \citep{dangel2020backpack}. Then, we calculate the mean stiffness of the value network across all state pairs and report its average computed over all training epochs. 

The green lines in \Cref{fig:stiffness_full} demonstrate that the stiffness decreases as the number of training levels increases in most of the Procgen games. This implies that a value network trained on more training environments is more likely to memorize training data and cannot extrapolate values of unseen states from the training environments.

We also conduct stiffness experiments for DCPG agents under the same experimental setting as PPG to evaluate how the delayed critic update mitigates the memorization problem. The blue lines in \Cref{fig:stiffness_full} show that DCPG achieves higher stiffness than PPG for all $n$ in 12 of 16 Procgen games. This suggests that the delayed critic update effectively alleviates the memorization problem.
 
\section{Training Value Function with Explicit Regularization}

We train PPG agents with discount regularization (PPG+DR) and activation regularization (PPG+AR) using 200 training levels for 25M environment steps on all 16 Procgen games and report the PPO-normalized training and test scores. For PPG+DR, we sweep the new discount factor $\gamma'$ in a range of $\{0.98, 0.99, 0.995\}$ and find $\gamma' = 0.995$ performs best. For PPG+AR, we sweep the regularization coefficient $\alpha$ within $\{0.01, 0.05, 0.1\}$ and find $\alpha = 0.05$ works best. \Cref{tab:procgen_regularization} shows that PPG+DR and PPG+AR improve both the training and test scores of PPG by 10\%p.

\begin{table}[h]
	\centering
	\caption{PPO-normalized training and test scores of PPG, PPG+DR, and PPG+AR. Each agent is trained on 200 training levels for 25M environment steps. The mean and standard deviation are computed over 10 different runs.}
    	\begin{tabular}{l|rrr}
	\toprule
	 & PPG & PPG+DR & PPG+AR  \\
	\midrule
	PPO-norm train score (\%) & 136.7 $\pm$ 4.4 & 146.8 $\pm$ 1.4 & 146.7 $\pm$ 2.6 \\
	PPO-norm test score (\%) & 160.3 $\pm$ 6.3 & 168.0 $\pm$ 5.8 & 169.9 $\pm$ 6.9 \\
	\bottomrule
	\end{tabular}
	\label{tab:procgen_regularization}
\end{table}

\section{Value Network Analysis of Delayed Critic Update}

To verify whether the delayed critic update acts as a value network regularizer, we train DCPG agents using 200 training levels for 25M environment steps on all 16 games of Procgen and compare the true and predicted values for the initial states of the training levels. More specifically, we collect 100 training episodes throughout the training and evaluate the value network prediction for the initial state of each trajectory. We estimate the true value of each initial state by computing the discounted return of the corresponding trajectory. We also conduct the same experiments with PPG+DR and PPG+AR for comparison with explicit value function regularization.

\begin{figure}[H]
    \centering
    \begin{tikzpicture}     	
 	\begin{groupplot}[
		group style = {group size = 4 by 4, horizontal sep=1.1cm, vertical sep=1.4cm},
    		width=4cm,
    		height=4cm,
    		grid=major,
    		xmin=0,
    		xmax=2.5e7,
    		scaled x ticks = false,
    		tick pos = left,
    		tick label style={font=\scriptsize},
    		xtick={0, 5e6, 10e6, 15e6, 20e6, 25e6},
    		xticklabels={0, 5, 10, 15, 20, 25},
		xlabel={Environment step ($10^6$)},
		ylabel={Value},
		label style={font=\scriptsize},
		ylabel style={at={(-0.15,0.5)}},
		xlabel style={at={(0.5,-0.15)}},
		title style={font=\small,anchor=north,yshift=8pt},
		legend style={font=\small,legend columns=2, /tikz/every even column/.append style={column sep=0.2cm}},
		legend to name=exp,
        ]
	\pgfplotsinvokeforeach{BigFish,BossFight,CaveFlyer,Chaser,Climber,CoinRun,Dodgeball,FruitBot,Heist,Jumper,Leaper,Maze,Miner,Ninja,Plunder,StarPilot}{
		\nextgroupplot[
			title=#1,
		]
        \addlegendimage{index of colormap=1 of Paired,line width=0.2mm}
        \addlegendentry{True}
        \addlegendimage{index of colormap=5 of Paired,line width=0.2mm}
        \addlegendentry{Predicted}        
    	\addplot+[index of colormap=1 of Paired,line width=0.2mm] table [x=num_steps, y=train_ret_n_disc_mean, col sep=comma]{data/progress-#1-ppg_norm_0.05.csv};
	\addplot+[index of colormap=5 of Paired,line width=0.2mm] table [x=num_steps, y=train_init_value_ests_mean, col sep=comma]{data/progress-#1-ppg_norm_0.05.csv};
	}
        \end{groupplot}	
        \node[above] at (current bounding box.north) {\pgfplotslegendfromname{exp}};
	\end{tikzpicture}
        	\caption{True and predicted values measured at the initial states of training environments for PPG with activation regularization (PPG+AR) on all 16 Procgen games. The mean is computed over 10 different runs.}
       	\label{fig:regularization_ppgar_full}
\end{figure}
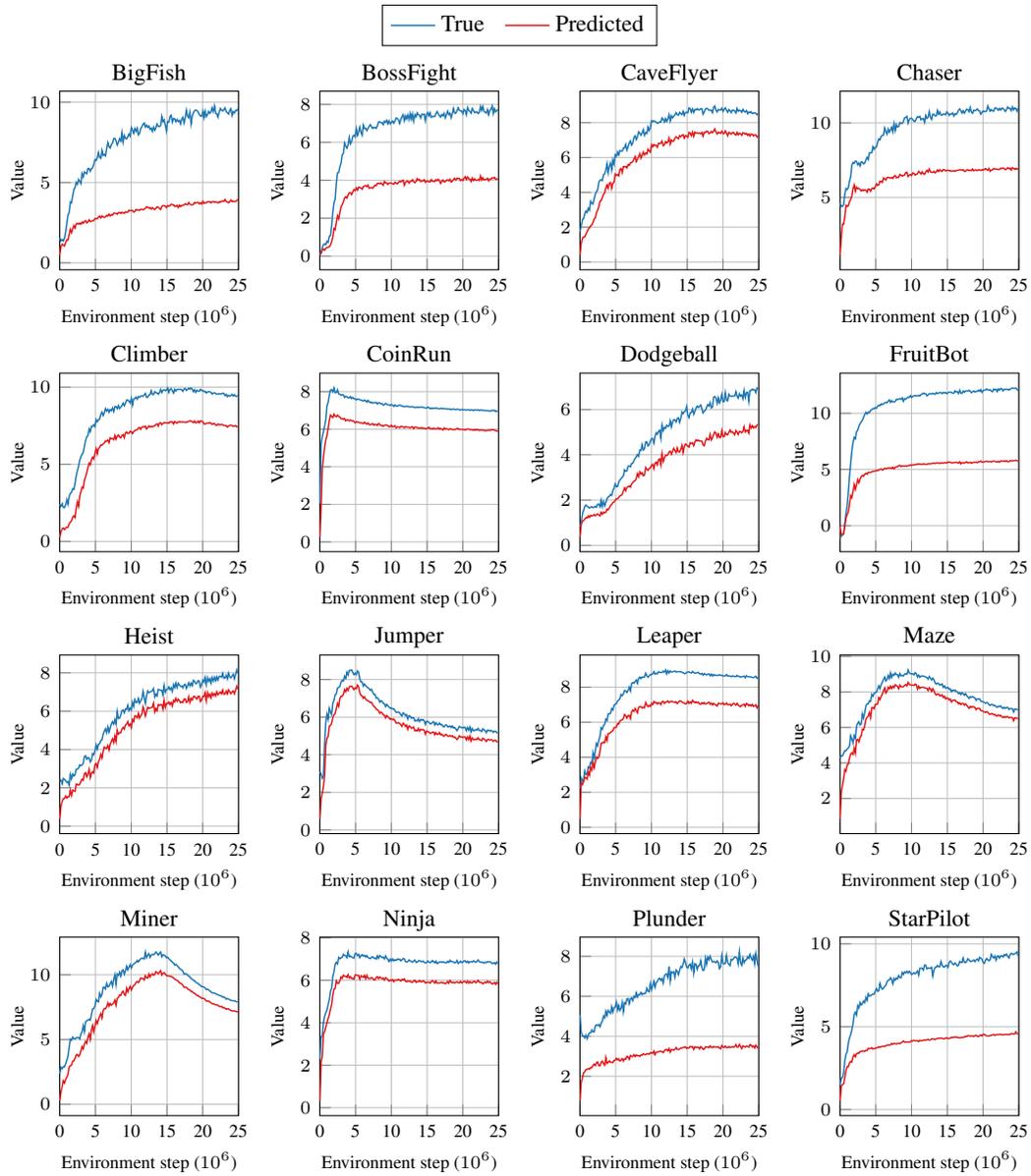

The red curves in \Cref{fig:regularization_dcpg_full} show that the value network of DCPG consistently underestimates the true values when the number of environment steps is small. In addition, the value predictions of DCPG become less biased as training progresses and reach the true values in most games. It implies that the delayed value update can serve as implicit regularization that slowly diminishes in strength, which can be beneficial both in the early and late stages of training. In contrast, the value network trained with discount and activation regularization fails to make unbiased predictions even at the end of the training, as shown in \Cref{fig:regularization_ppgdr_full,fig:regularization_ppgar_full}.

\begin{figure}[H]
    \centering
    \begin{tikzpicture}
        \begin{groupplot}[
		group style = {group name=valuegrad, group size = 4 by 4, horizontal sep=1.1cm, vertical sep=1.4cm},
		width=4.0cm,
		height=4.0cm,
		grid=major,
		xmin=1,
		xmax=200,
		scaled x ticks = false,
		scaled y ticks = false,
		tick pos = left,
		tick label style={font=\scriptsize},
		xtick={1, 2, 10, 50, 200},
		xticklabels={1, 2, 10, 50, 200},
		minor tick style={draw=none},
		y tick label style={/pgf/number format/fixed},
		max space between ticks=1000pt,
		try min ticks=4,
		xlabel={\# training levels},
		ylabel={Stiffness},
		label style={font=\scriptsize},
		ylabel style={at={(-0.2,0.5)}},
		xlabel style={at={(0.5,-0.15)}},
		title style={font=\small,anchor=north,yshift=8pt},
		legend style={legend columns=5, font=\small, /tikz/every even column/.append style={column sep=0.2cm}},
		legend to name=exp,
        ]
		\pgfplotsinvokeforeach{BigFish,BossFight,CaveFlyer,Chaser,Climber,CoinRun,Dodgeball,FruitBot,Heist,Jumper,Leaper,Maze,Miner,Ninja,Plunder,StarPilot}{
		    \nextgroupplot[
		        title=#1,
		        xmode=log
		    ]
    		\addplot+[index of colormap=3 of Paired,line width=0.30mm,mark=*,mark options={scale=0.5}] table [x=num_train_envs, y=#1_value_grad, col sep=comma]{data/ppg_grad_avg.csv};
		\addlegendentry{PPG}
    		\addplot+[index of colormap=1 of Paired,line width=0.30mm,mark=*,mark options={scale=0.5}] table [x=num_train_envs, y=#1_value_grad, col sep=comma]{data/appg_grad_avg.csv};
		\addlegendentry{DCPG}
		}
        \end{groupplot}
	\node[above] at (current bounding box.north) {\pgfplotslegendfromname{exp}};
    \end{tikzpicture}
    \caption{Average stiffness of value networks for PPG and DCPG on all 16 Procgen games while varying the number of training levels. Each agent is trained for 8M environment steps.}
       \label{fig:stiffness_full}
\end{figure}
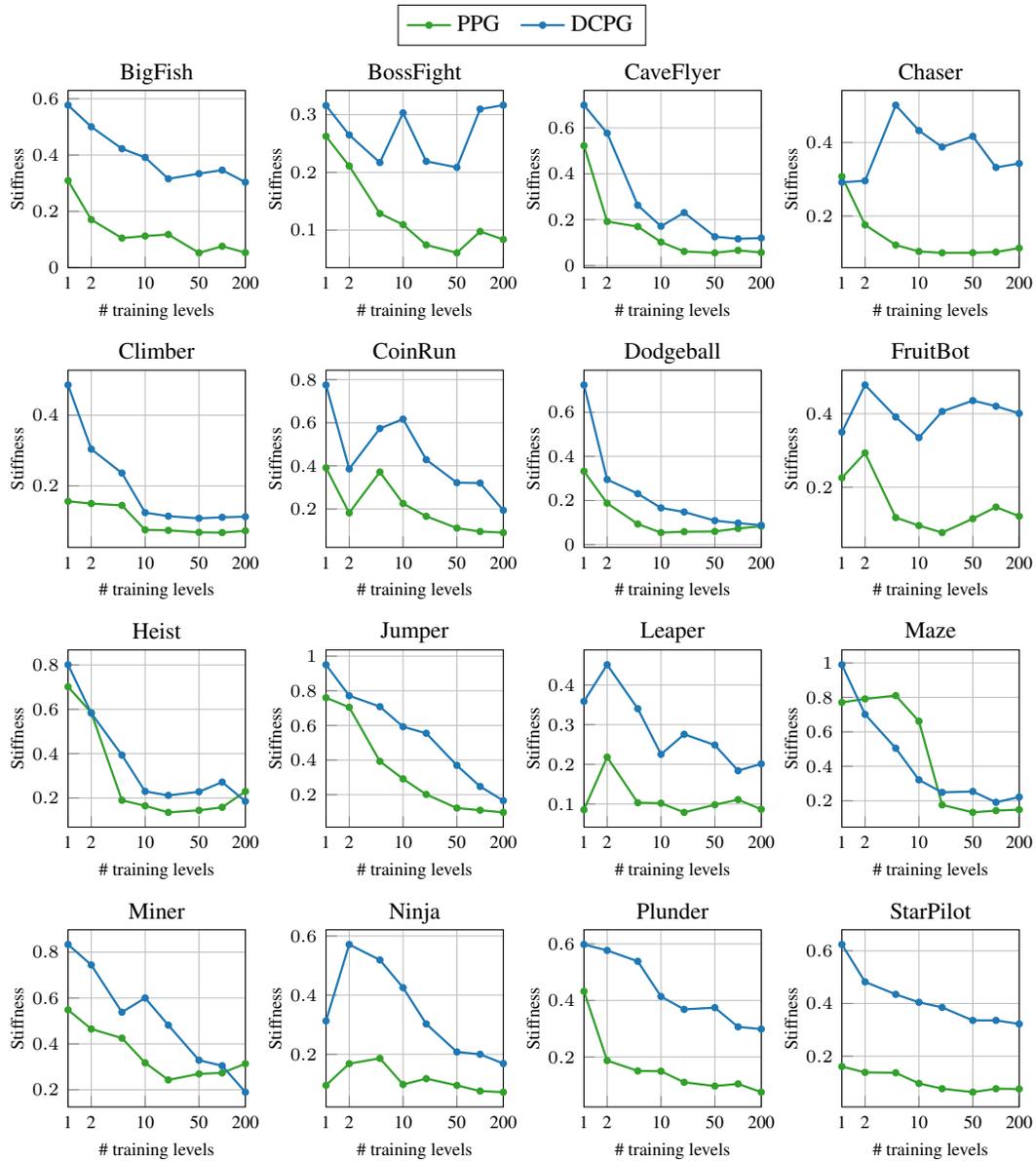 

\begin{figure}[H]
    \centering
    \begin{tikzpicture}     	
 	\begin{groupplot}[
		group style = {group size = 4 by 4, horizontal sep=1.1cm, vertical sep=1.4cm},
    		width=4cm,
    		height=4cm,
    		grid=major,
    		xmin=0,
    		xmax=2.5e7,
    		scaled x ticks = false,
    		tick pos = left,
    		tick label style={font=\scriptsize},
    		xtick={0, 5e6, 10e6, 15e6, 20e6, 25e6},
    		xticklabels={0, 5, 10, 15, 20, 25},
		xlabel={Environment step ($10^6$)},
		ylabel={Value},
		label style={font=\scriptsize},
		ylabel style={at={(-0.15,0.5)}},
		xlabel style={at={(0.5, -0.15)}},
		title style={font=\small,anchor=north,yshift=8pt},
		legend style={font=\small,legend columns=2, /tikz/every even column/.append style={column sep=0.2cm}},
		legend to name=exp,
        ]
	\pgfplotsinvokeforeach{BigFish,BossFight,CaveFlyer,Chaser,Climber,CoinRun,Dodgeball,FruitBot,Heist,Jumper,Leaper,Maze,Miner,Ninja,Plunder,StarPilot}{
		\nextgroupplot[
			title=#1,
		]
        \addlegendimage{index of colormap=1 of Paired,line width=0.2mm}
        \addlegendentry{True}
        \addlegendimage{index of colormap=5 of Paired,line width=0.2mm}
        \addlegendentry{Predicted}        
    	\addplot+[index of colormap=1 of Paired,line width=0.2mm] table [x=num_steps, y=train_ret_n_disc_mean, col sep=comma]{data/progress-#1-appg.csv};
	\addplot+[index of colormap=5 of Paired,line width=0.2mm] table [x=num_steps, y=train_init_value_ests_mean, col sep=comma]{data/progress-#1-appg.csv};
	}
        \end{groupplot}	
        \node[above] at (current bounding box.north) {\pgfplotslegendfromname{exp}};
	\end{tikzpicture}
        	\caption{True and predicted values measured at the initial states of training environments for DCPG on all 16 Procgen games. The mean is computed over 10 different runs.}
       	\label{fig:regularization_dcpg_full}
\end{figure}
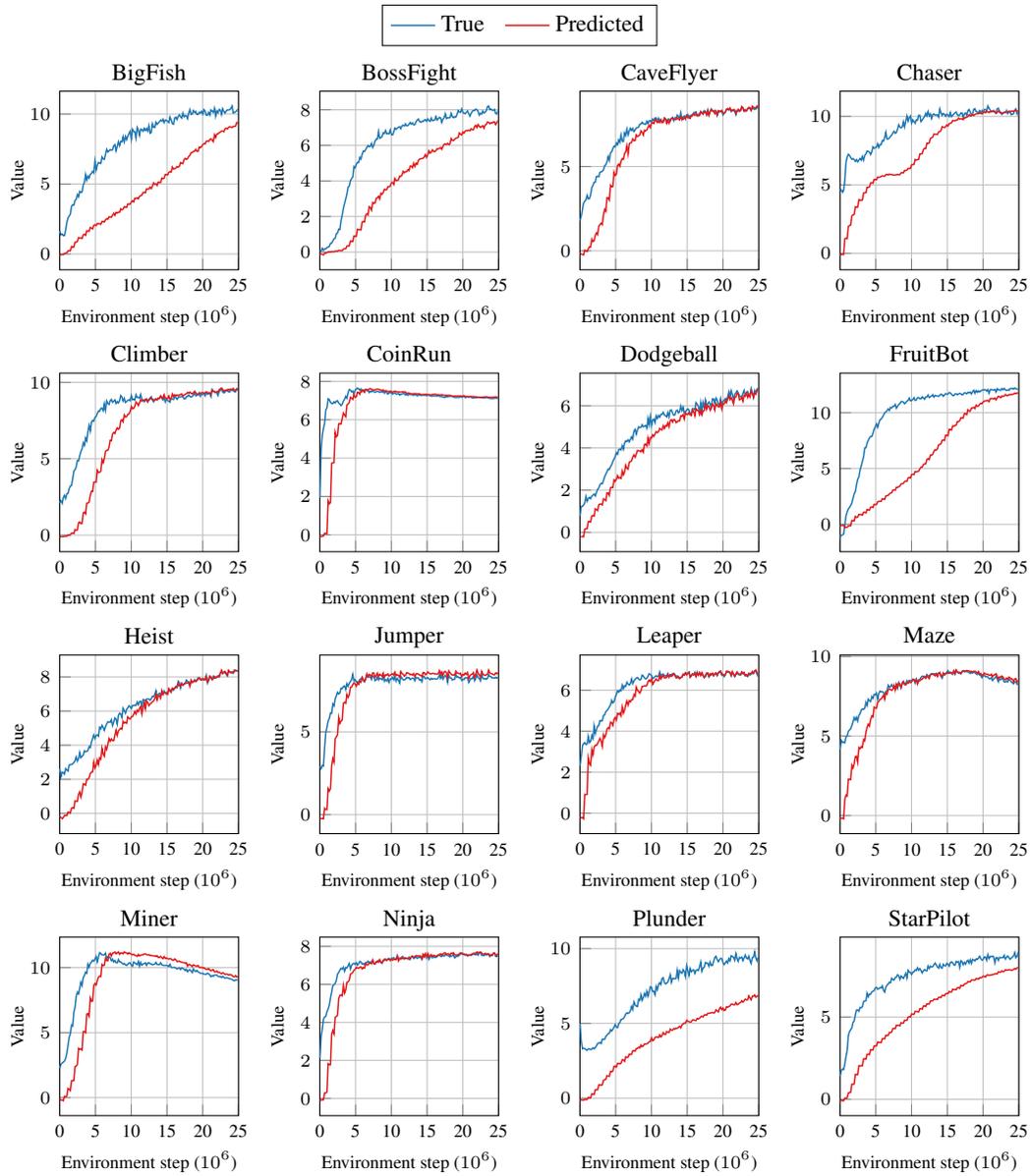

\begin{figure}[H]
    \centering
    \begin{tikzpicture}     	
 	\begin{groupplot}[
		group style = {group size = 4 by 4, horizontal sep=1.1cm, vertical sep=1.4cm},
    		width=4cm,
    		height=4cm,
    		grid=major,
    		xmin=0,
    		xmax=2.5e7,
    		scaled x ticks = false,
    		tick pos = left,
    		tick label style={font=\scriptsize},
    		xtick={0, 5e6, 10e6, 15e6, 20e6, 25e6},
    		xticklabels={0, 5, 10, 15, 20, 25},
		xlabel={Environment step ($10^6$)},
		ylabel={Value},
		label style={font=\scriptsize},
		ylabel style={at={(-0.15,0.5)}},
		xlabel style={at={(0.5,-0.15)}},
		title style={font=\small,anchor=north,yshift=8pt},
		legend style={font=\small,legend columns=2, /tikz/every even column/.append style={column sep=0.2cm}},
		legend to name=exp,
        ]
	\pgfplotsinvokeforeach{BigFish,BossFight,CaveFlyer,Chaser,Climber,CoinRun,Dodgeball,FruitBot,Heist,Jumper,Leaper,Maze,Miner,Ninja,Plunder,StarPilot}{
		\nextgroupplot[
			title=#1,
		]
        \addlegendimage{index of colormap=1 of Paired,line width=0.2mm}
        \addlegendentry{True}
        \addlegendimage{index of colormap=5 of Paired,line width=0.2mm}
        \addlegendentry{Predicted}        
    	\addplot+[index of colormap=1 of Paired,line width=0.2mm] table [x=num_steps, y=train_ret_n_disc_mean, col sep=comma]{data/progress-#1-ppg_discount_0.995.csv};
	\addplot+[index of colormap=5 of Paired,line width=0.2mm] table [x=num_steps, y=train_init_value_ests_mean, col sep=comma]{data/progress-#1-ppg_discount_0.995.csv};
	}
        \end{groupplot}	
        \node[above] at (current bounding box.north) {\pgfplotslegendfromname{exp}};
	\end{tikzpicture}
        	\caption{True and predicted values measured at the initial states of training environments for PPG with discount regularization (PPG+DR) on all 16 Procgen games. The mean is computed over 10 different runs.}
       	\label{fig:regularization_ppgdr_full}
\end{figure}
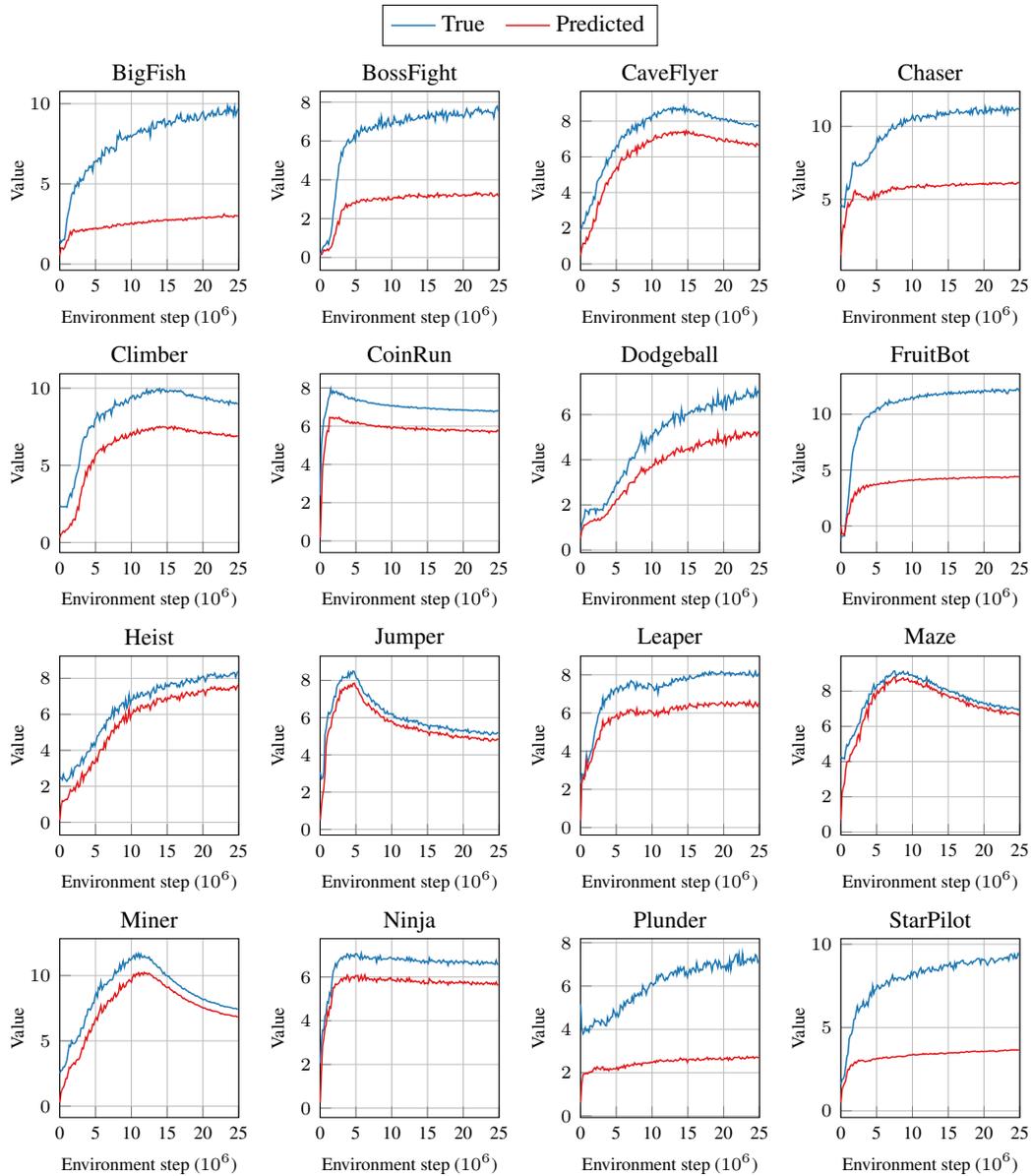

\newpage

\section{Implementation Details \& Hyperparameters}

For all experiments, we implement policy and value networks using the ResNet architecture proposed in IMPALA \cite{espeholt2018impala} and train the networks using Adam optimizer \citep{kingma2014adam}, following the standard practice in the prior works \citep{cobbe2020leveraging,cobbe2021phasic}. We conduct all experiments using Intel Xeon Gold 5220R CPU, 64GB RAM, and NVIDIA RTX 2080 Ti GPU. We use PyTorch as a deep learning framework \citep{paszke2019pytorch}.

\paragraph{PPO} 

We use the implementation of PPO by \citet{pytorchrl}. Unless otherwise specified, we use the same hyperparameter setting provided in \citet{cobbe2020leveraging} to reproduce PPO and PPO-based methods. The values of the hyperparameters used are shown in \Cref{tab:ppo_hp}.

\begin{table}[h]
    \centering
    \caption{PPO hyperparameters.}
    \label{tab:ppo_hp}
    \begin{tabular}{lc}
    \toprule
    Hyperparameter & Value \\
    \midrule
    Discount factor ($\gamma$) & 0.999 \\
    GAE smoothing parameter ($\lambda$) & 0.95 \\
    \# timesteps per rollout & 256 \\
    \# epochs per rollout & 3 \\
    \# minibatches per epoch & 8 \\
    Entropy bonus & 0.01 \\
    PPO clip range ($\epsilon$) & 0.2 \\
    Reward normalization? & Yes \\
    Learning rate & 5e-4 \\
    \# workers & 1 \\
    \# environments per worker & 64 \\
    Total timesteps & 25M \\
    LSTM? & No \\
    Frame stack? & No \\
    \bottomrule
    \end{tabular}
    \label{tab:ppo_hp}
\end{table}

\paragraph{UCB-DrAC}

We use the official implementation by the authors\footnote{\url{https://github.com/rraileanu/auto-drac}}. We use the best hyperparameter setting provided in the original paper. More specifically, we use regularization coefficient $\alpha_r=0.1$, exploration coefficient $c=0.1$, and sliding window size $K=10$ for all Procgen games.

\paragraph{PLR}

We reproduce the reported results of PLR using the implementation released by the authors\footnote{\url{https://github.com/facebookresearch/level-replay}}. We use the recommended hyperparameter setting presented in the original paper and use $L_1$ value loss as the scoring function, rank prioritization, temperature $\beta=0.1$, and staleness coefficient $\rho=0.1$ for all Procgen games.

\paragraph{PPG} 

We build PPG on top of the implementation of PPO. For PPO hyperparameters, we use the default hyperparameter setting in \Cref{tab:ppo_hp}, except for the number of PPO epochs. For PPG-specific hyperparameters, we use the best hyperparameter setting provided in \citet{cobbe2021phasic}. The values of the PPG-specific hyperparameter are shown in \Cref{tab:ppg_hp2}. The policy regularizer coefficient $\beta_\pi$ denotes the weight for the policy regularizer $C_\pi$ when jointly optimized with the auxiliary objective $J_\mathrm{aux}$. We also provide the pseudocode of PPG in \Cref{alg:ppg}.

\begin{algorithm}[h]
	\small
	\caption{Phasic Policy Gradient (PPG)}
	\begin{algorithmic}[1]
	\Require Policy network $\pi_\theta$, value network $V_\phi$, auxiliary value head $V_\theta$
    	\For{phase = $1, 2, \ldots $}
    	\State Initialize buffer $\mathcal{B}$
    	\For{iter = $1, 2, \ldots, N_\pi$} \Comment{Policy phase}
    	\State Sample trajectories $\tau$ using $\pi_\theta$ and compute value function target $\hat{R}_t$ for each state $s_t \in \tau$
    	\For{epoch = $1, 2, \ldots, E_\pi$}
    	\State Optimize policy objective $J_\pi(\theta)$ and value objective $J_V(\phi)$ with $\tau$
    	\EndFor
	\State Add $(s_t, \hat{R}_t)$ to $\mathcal{B}$
    	\EndFor
	\For{iter = $1, 2, \ldots, E_\mathrm{aux}$} \Comment{Auxiliary phase}
	\State Optimize value objective $J_V(\phi)$, auxiliary objective $J_\mathrm{aux}(\theta)$, and policy regularizer $C_\pi(\theta)$ with $\mathcal{B}$
	\EndFor
    	\EndFor
	\end{algorithmic}
	\label{alg:ppg}
\end{algorithm}

\begin{table}[h]
    \centering
    \caption{PPG-specific hyperparameters.}
    \begin{tabular}{lc}
    \toprule
    Hyperparameter & Value \\
    \midrule
    \# policy phases per auxiliary phase ($N_\pi$) & 32 \\
    \# policy epochs ($E_\pi$) & 1 \\
    \# value epochs ($E_V$) & 1 \\
    \# auxiliary epochs ($E_\mathrm{aux}$) & 6 \\
    Policy regularizer coefficient ($\beta_\pi$)& 1.0 \\
    \# minibatches per auxiliary epoch per $N_\pi$ & 16 \\
    \bottomrule
    \end{tabular}
    \label{tab:ppg_hp2}
\end{table}

\paragraph{DAAC}

For DAAC and IDAAC, which adds an auxiliary regularizer to the DAAC policy network, we use the official code released by the authors\footnote{\url{https://github.com/rraileanu/idaac}}. We find that some hyperparameters should be set differently for each Procgen game to reproduce the results reported in the original paper. For a fair comparison with other methods, we use the same set of hyperparameters for all Procgen games with the best overall performance provided by the authors. More specifically, we use $E_\pi=1$, $E_V=9$, $N_\pi=1$, $\alpha_a=0.25$, and $\alpha_i=0.001$. We also find that the performance of DAAC is better than IDAAC when using a single set of hyperparameters, as shown in \Cref{tab:procgen_daac}. Therefore, we compare our methods with DAAC. 

\begin{table}[H]
	\centering
	\caption{PPO-normalized train and test scores of DAAC and IDAAC. Each agent is trained on 200 training levels for 25M environment steps. The mean and standard deviation are computed over 10 different runs.}
    	\begin{tabular}{l|rr}
	\toprule
	 & DAAC & IDAAC \\
	\midrule
	PPO-norm train score (\%) & 104.4 $\pm$ 4.1 & 99.7 $\pm$ 5.4 \\
	PPO-norm test score (\%) & 136.7 $\pm$ 7.6 & 129.9 $\pm$ 7.8 \\
	\bottomrule
	\end{tabular}
    	\label{tab:procgen_daac}
\end{table}

\paragraph{DCPG} 

We build DCPG on top of the implementation of PPG. DCPG introduces one additional hyperparameter compared to PPG, named the value regularizer coefficient. The value regularizer coefficient $\beta_V$ denotes the weight for the value regularizer $C_V$ when jointly optimized with the policy objective $J_\pi$. We use the default hyperparameter setting in \Cref{tab:ppg_hp2} for PPG hyperparameters. We set the value regularizer coefficient to $\beta_V = 1.0$ without any hyperparameter tuning. 

\paragraph{DDCPG}

We implement the discriminator for the dynamics learning using a fully-connected layer of hidden sizes $[256, 256]$ with ReLU activation. DDCPG introduces two additional hyperparameters compared to DCPG. First, the dynamics objective coefficient $\beta_f$ denotes the weight of the dynamics objective $J_f$ when jointly optimized with the value objective $J_V$ and the policy regularizer $C_\pi$. We set the dynamics objective coefficient to $\beta_f = 1.0$ without any hyperparameter tuning. Second, the inverse dynamics coefficient $\eta$ is the hyperparameter that controls the strength of the inverse dynamics learning relative to the forward dynamics learning. We sweep the inverse dynamics coefficient within a range of $\eta \in \{ 0.5, 1.0 \}$ and choose $\eta=0.5$. We use the same hyperparameter setting as DCPG for the other hyperparameters.

\newpage

\section{More Results on Procgen Benchmark}

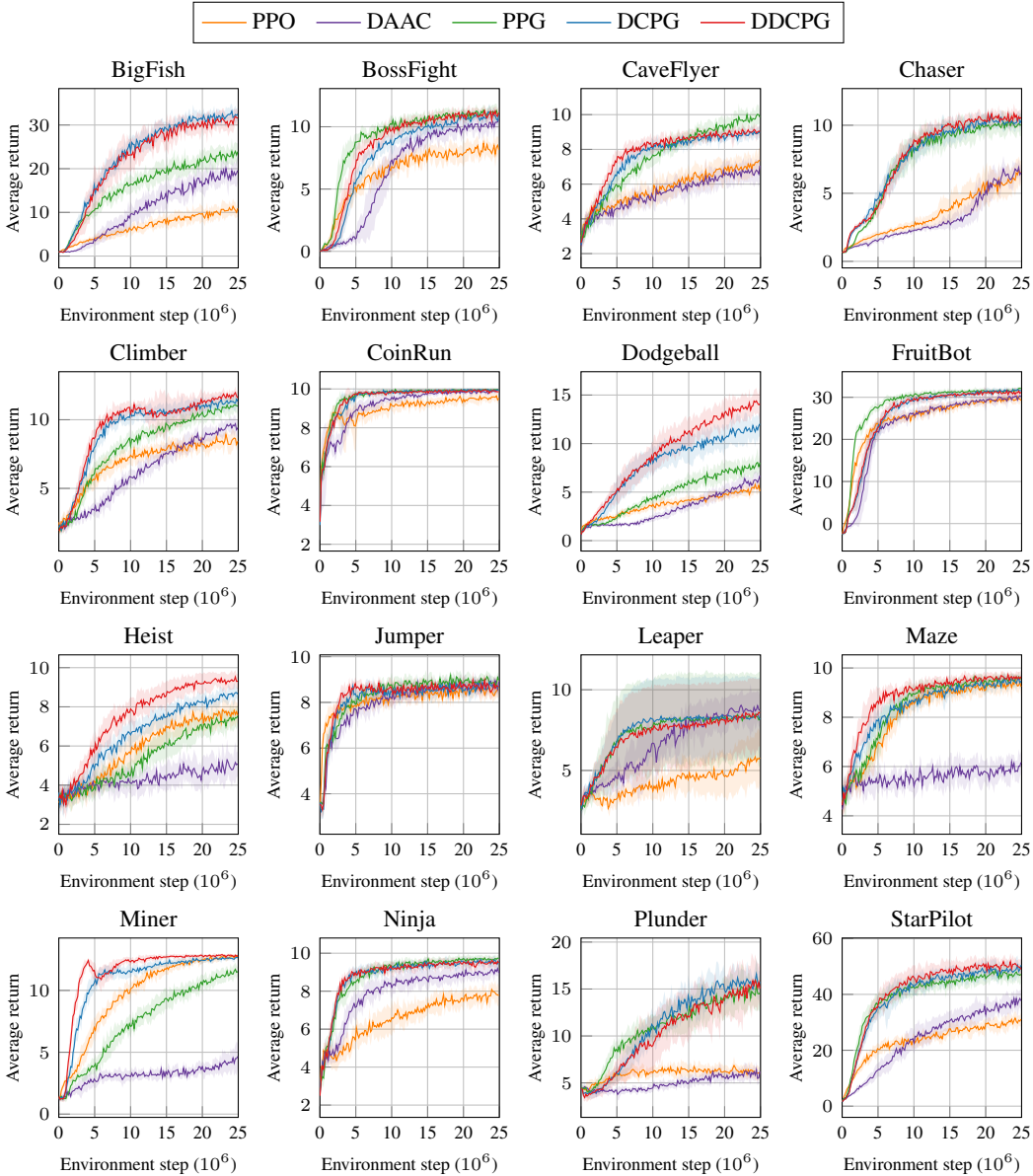
\begin{figure}[H]
    \centering
    \begin{tikzpicture}
        \begin{groupplot}[
		group style = {group name=procgen, group size = 4 by 4, horizontal sep=1.1cm, vertical sep=1.4cm},
		width=4.0cm,
		height=4.0cm,
		grid=major,
		xmin=0,
		xmax=2.5e7,
		scaled x ticks = false,
		tick pos = left,
		tick label style={font=\scriptsize},
		xtick={0, 5e6, 10e6, 15e6, 20e6, 25e6},
		xticklabels={0, 5, 10, 15, 20, 25},
		xlabel={Environment step ($10^6$)},
		ylabel={Average return},
		label style={font=\scriptsize},
		ylabel style={at={(-0.15,0.5)}},
		xlabel style={at={(0.5,-0.15)}},
		title style={font=\small,anchor=north,yshift=8pt},
		legend style={legend columns=5, font=\small, /tikz/every even column/.append style={column sep=0.2cm}},
		legend to name=exp,
        ]
	\pgfplotsinvokeforeach{BigFish,BossFight,CaveFlyer,Chaser,Climber,CoinRun,Dodgeball,FruitBot,Heist,Jumper,Leaper,Maze,Miner,Ninja,Plunder,StarPilot}{
		\nextgroupplot[
			title=#1,
		]
		\addlegendimage{index of colormap=7 of Paired,line width=0.2mm}
		\addlegendentry{PPO}
		\addlegendimage{index of colormap=9 of Paired,line width=0.2mm}
    		\addlegendentry{DAAC}
		\addlegendimage{index of colormap=3 of Paired,line width=0.2mm}
    		\addlegendentry{PPG}
		\addlegendimage{index of colormap=1 of Paired,line width=0.2mm}
    		\addlegendentry{DCPG}
		\addlegendimage{index of colormap=5 of Paired,line width=0.2mm}
    		\addlegendentry{DDCPG}
    		\addplot+[index of colormap=7 of Paired,line width=0.15mm] table [x=num_steps, y=train_episode_rewards_mean, col sep=comma]{data/progress-#1-ppo.csv};
    		\addplot+[index of colormap=9 of Paired,line width=0.15mm] table [x=num_steps, y=train_episode_rewards_mean, col sep=comma]{data/progress-#1-daac.csv};
    		\addplot+[index of colormap=3 of Paired,line width=0.15mm] table [x=num_steps, y=train_episode_rewards_mean, col sep=comma]{data/progress-#1-ppg.csv};
    		\addplot+[index of colormap=1 of Paired,line width=0.15mm] table [x=num_steps, y=train_episode_rewards_mean, col sep=comma]{data/progress-#1-appg.csv};
    		\addplot+[index of colormap=5 of Paired,line width=0.15mm] table [x=num_steps, y=train_episode_rewards_mean, col sep=comma]{data/progress-#1-ssl_v1.csv};
    		\addplot[name path=plus,draw=none] table [x=num_steps, y=train_episode_rewards_mean_plus_std, col sep=comma]{data/progress-#1-ppo.csv};
    		\addplot[name path=minus,draw=none] table [x=num_steps, y=train_episode_rewards_mean_minus_std, col sep=comma]{data/progress-#1-ppo.csv};
    		\addplot[index of colormap=7 of Paired,fill opacity=0.12] fill between[of=plus and minus];
    		\addplot[name path=plus,draw=none] table [x=num_steps, y=train_episode_rewards_mean_plus_std, col sep=comma]{data/progress-#1-daac.csv};
    		\addplot[name path=minus,draw=none] table [x=num_steps, y=train_episode_rewards_mean_minus_std, col sep=comma]{data/progress-#1-daac.csv};
    		\addplot[index of colormap=9 of Paired,fill opacity=0.12] fill between[of=plus and minus];
    		\addplot[name path=plus,draw=none] table [x=num_steps, y=train_episode_rewards_mean_plus_std, col sep=comma]{data/progress-#1-ppg.csv};
    		\addplot[name path=minus,draw=none] table [x=num_steps, y=train_episode_rewards_mean_minus_std, col sep=comma]{data/progress-#1-ppg.csv};
    		\addplot[index of colormap=3 of Paired,fill opacity=0.12] fill between[of=plus and minus];
    		\addplot[name path=plus,draw=none] table [x=num_steps, y=train_episode_rewards_mean_plus_std, col sep=comma]{data/progress-#1-appg.csv};
    		\addplot[name path=minus,draw=none] table [x=num_steps, y=train_episode_rewards_mean_minus_std, col sep=comma]{data/progress-#1-appg.csv};
    		\addplot[index of colormap=1 of Paired,fill opacity=0.12] fill between[of=plus and minus];
    		\addplot[name path=plus,draw=none] table [x=num_steps, y=train_episode_rewards_mean_plus_std, col sep=comma]{data/progress-#1-ssl_v1.csv};
    		\addplot[name path=minus,draw=none] table [x=num_steps, y=train_episode_rewards_mean_minus_std, col sep=comma]{data/progress-#1-ssl_v1.csv};
    		\addplot[index of colormap=5 of Paired,fill opacity=0.12] fill between[of=plus and minus];
		}
        \end{groupplot}
	\node[above] at (current bounding box.north) {\pgfplotslegendfromname{exp}};
    \end{tikzpicture}
    \caption{Training performance curves of each method on all 16 Procgen games. Each agent is trained on 200 training levels for 25M environment steps and evaluated on the same training levels. The mean and standard deviation are computed over 10 different runs.}
    \label{fig:procgen_train_full}
\end{figure}

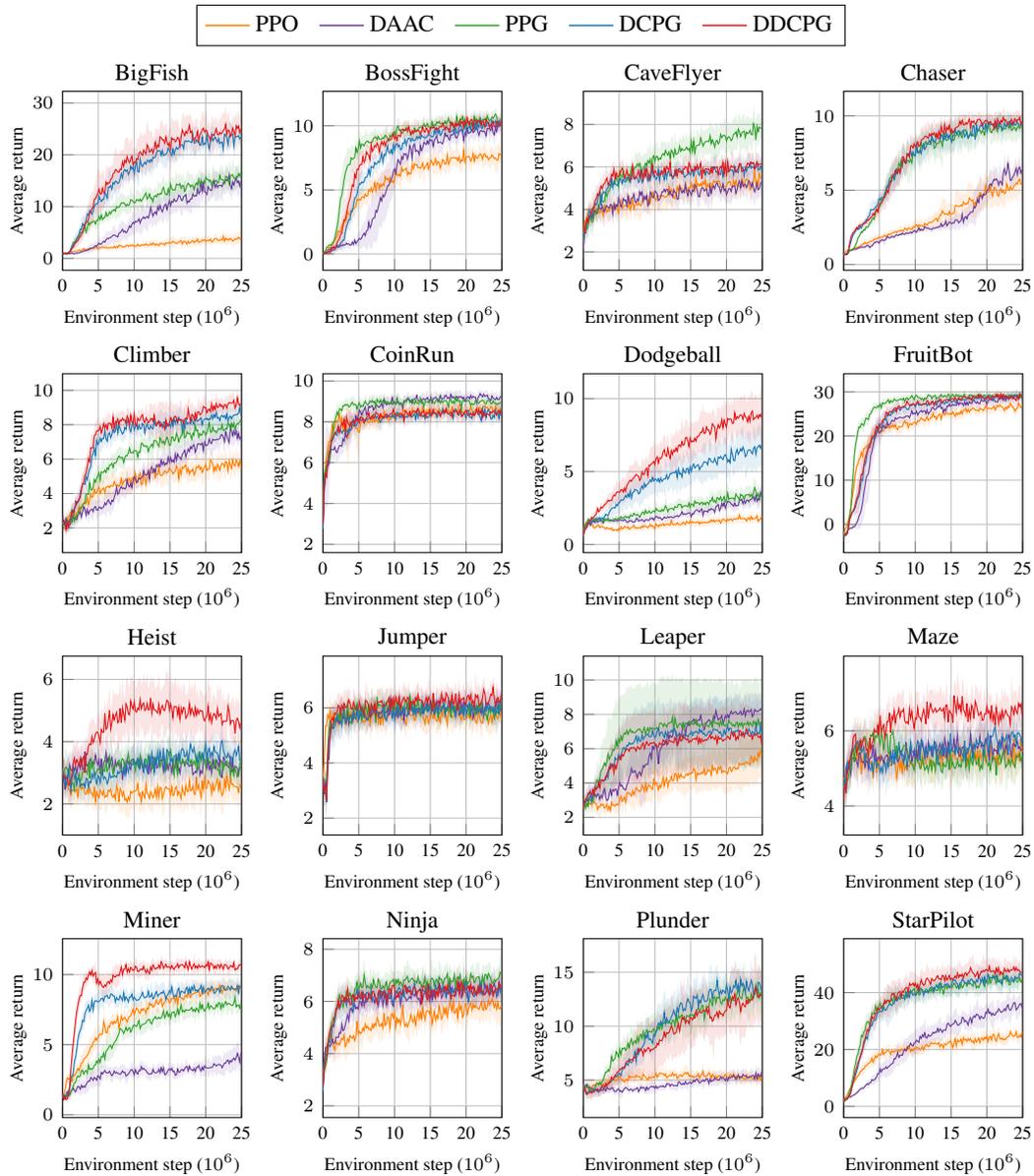
\begin{figure}[H]
    \centering
    \begin{tikzpicture}
        \begin{groupplot}[
		group style = {group name=procgen, group size = 4 by 4, horizontal sep=1.1cm, vertical sep=1.4cm},
		width=4.0cm,
		height=4.0cm,
		grid=major,
		xmin=0,
		xmax=2.5e7,
		scaled x ticks = false,
		tick pos = left,
		tick label style={font=\scriptsize},
		xtick={0, 5e6, 10e6, 15e6, 20e6, 25e6},
		xticklabels={0, 5, 10, 15, 20, 25},
		xlabel={Environment step ($10^6$)},
		ylabel={Average return},
		label style={font=\scriptsize},
		ylabel style={at={(-0.15,0.5)}},
		xlabel style={at={(0.5,-0.15)}},
		title style={font=\small,anchor=north,yshift=8pt},
		legend style={legend columns=5, font=\small, /tikz/every even column/.append style={column sep=0.2cm}},
		legend to name=exp,
        ]
	\pgfplotsinvokeforeach{BigFish,BossFight,CaveFlyer,Chaser,Climber,CoinRun,Dodgeball,FruitBot,Heist,Jumper,Leaper,Maze,Miner,Ninja,Plunder,StarPilot}{
		\nextgroupplot[
			title=#1,
		]
		\addlegendimage{index of colormap=7 of Paired,line width=0.2mm}
		\addlegendentry{PPO}
		\addlegendimage{index of colormap=9 of Paired,line width=0.2mm}
    		\addlegendentry{DAAC}
		\addlegendimage{index of colormap=3 of Paired,line width=0.2mm}
    		\addlegendentry{PPG}
		\addlegendimage{index of colormap=1 of Paired,line width=0.2mm}
    		\addlegendentry{DCPG}
		\addlegendimage{index of colormap=5 of Paired,line width=0.2mm}
    		\addlegendentry{DDCPG}
    		\addplot+[index of colormap=7 of Paired,line width=0.15mm] table [x=num_steps, y=test_episode_rewards_mean, col sep=comma]{data/progress-#1-ppo.csv};
    		\addplot+[index of colormap=9 of Paired,line width=0.15mm] table [x=num_steps, y=test_episode_rewards_mean, col sep=comma]{data/progress-#1-daac.csv};
    		\addplot+[index of colormap=3 of Paired,line width=0.15mm] table [x=num_steps, y=test_episode_rewards_mean, col sep=comma]{data/progress-#1-ppg.csv};
    		\addplot+[index of colormap=1 of Paired,line width=0.15mm] table [x=num_steps, y=test_episode_rewards_mean, col sep=comma]{data/progress-#1-appg.csv};
    		\addplot+[index of colormap=5 of Paired,line width=0.15mm] table [x=num_steps, y=test_episode_rewards_mean, col sep=comma]{data/progress-#1-ssl_v1.csv};
    		\addplot[name path=plus,draw=none] table [x=num_steps, y=test_episode_rewards_mean_plus_std, col sep=comma]{data/progress-#1-ppo.csv};
    		\addplot[name path=minus,draw=none] table [x=num_steps, y=test_episode_rewards_mean_minus_std, col sep=comma]{data/progress-#1-ppo.csv};
    		\addplot[index of colormap=7 of Paired,fill opacity=0.12] fill between[of=plus and minus];
    		\addplot[name path=plus,draw=none] table [x=num_steps, y=test_episode_rewards_mean_plus_std, col sep=comma]{data/progress-#1-daac.csv};
    		\addplot[name path=minus,draw=none] table [x=num_steps, y=test_episode_rewards_mean_minus_std, col sep=comma]{data/progress-#1-daac.csv};
    		\addplot[index of colormap=9 of Paired,fill opacity=0.12] fill between[of=plus and minus];
    		\addplot[name path=plus,draw=none] table [x=num_steps, y=test_episode_rewards_mean_plus_std, col sep=comma]{data/progress-#1-ppg.csv};
    		\addplot[name path=minus,draw=none] table [x=num_steps, y=test_episode_rewards_mean_minus_std, col sep=comma]{data/progress-#1-ppg.csv};
    		\addplot[index of colormap=3 of Paired,fill opacity=0.12] fill between[of=plus and minus];
    		\addplot[name path=plus,draw=none] table [x=num_steps, y=test_episode_rewards_mean_plus_std, col sep=comma]{data/progress-#1-appg.csv};
    		\addplot[name path=minus,draw=none] table [x=num_steps, y=test_episode_rewards_mean_minus_std, col sep=comma]{data/progress-#1-appg.csv};
    		\addplot[index of colormap=1 of Paired,fill opacity=0.12] fill between[of=plus and minus];
    		\addplot[name path=plus,draw=none] table [x=num_steps, y=test_episode_rewards_mean_plus_std, col sep=comma]{data/progress-#1-ssl_v1.csv};
    		\addplot[name path=minus,draw=none] table [x=num_steps, y=test_episode_rewards_mean_minus_std, col sep=comma]{data/progress-#1-ssl_v1.csv};
    		\addplot[index of colormap=5 of Paired,fill opacity=0.12] fill between[of=plus and minus];
		}
        \end{groupplot}
	\node[above] at (current bounding box.north) {\pgfplotslegendfromname{exp}};
    \end{tikzpicture}
    \caption{Test performance curves of each method on all 16 Procgen games. Each agent is trained on 200 training levels for 25M environment steps and evaluated on 100 unseen test levels. The mean and standard deviation are computed over 10 different runs.}
    \label{fig:procgen_test_full}
\end{figure}

The training and test performance curves of DCPG, DDCPG, and baseline methods for all 16 Procgen games are shown in \Cref{fig:procgen_train_full,fig:procgen_test_full}, respectively. The results of UCB-DrAC and PLR are omitted in the figures for better visibility. The average training returns of DCPG, DDCPG, and all baselines are presented in \Cref{tab:procgen_train}. Our methods also achieve better training performance and sample efficiency than all baselines.

\newpage

\begin{table}[H]
    \centering
    \caption{Average training returns of each method on all 16 Procgen games. Each agent is trained on 200 training levels for 25M environment steps and evaluated on the same training levels. The mean and standard deviation are computed over 10 different runs.}
    \label{tab:procgen_train}
    \adjustbox{max width=\textwidth}{
    \begin{tabular}{l|rrrrr|rr}
        \toprule
        Environment & PPO & UCB-DrAC & PLR & DAAC & PPG & DCPG & DDCPG \\
        \midrule
        BigFish & 11.0 $\pm$ 1.7 & 12.2 $\pm$ 3.1 & 13.6 $\pm$ 3.0 & 18.4 $\pm$ 3.3 & 24.1 $\pm$ 1.6 & {32.8} $\pm$ {2.0} & {31.8} $\pm$ {2.1} \\
        BossFight & 7.9 $\pm$ 0.7 & 8.2 $\pm$ 0.7 & 8.8 $\pm$ 0.7 & 9.9 $\pm$ 0.7 & {11.1} $\pm$ {0.3} & 10.7 $\pm$ 0.6 & 11.0 $\pm$ 0.5 \\
        CaveFlyer & 7.1 $\pm$ 0.9 & 6.0 $\pm$ 0.9 & 7.3 $\pm$ 0.5 & 6.8 $\pm$ 0.8 & {10.0} $\pm$ {0.5} & 9.0 $\pm$ 0.5 & 9.0 $\pm$ 0.3 \\
        Chaser & 6.4 $\pm$ 0.8 & 7.3 $\pm$ 0.7 & 8.0 $\pm$ 0.6 & 6.4 $\pm$ 1.0 & 10.0 $\pm$ 0.7 & {10.4} $\pm$ {0.6} & {10.6} $\pm$ {0.5} \\
        Climber & 8.4 $\pm$ 0.5 & 8.4 $\pm$ 0.6 & 8.6 $\pm$ 0.6 & 9.3 $\pm$ 0.6 & 10.9 $\pm$ 0.4 & {11.3} $\pm$ {0.5} & {11.9} $\pm$ {0.4}  \\
        CoinRun & 9.6 $\pm$ 0.1 & 9.5 $\pm$ 0.2 & 9.4 $\pm$ 0.3 & 9.9 $\pm$ 0.1 & {10.0} $\pm$ {0.1} & 9.9 $\pm$ 0.1 & 9.9 $\pm$ 0.1 \\
        Dodgeball & 5.5 $\pm$ 0.7 & 8.0 $\pm$ 1.2 & 5.1 $\pm$ 0.8 & 6.4 $\pm$ 0.9 & 7.9 $\pm$ 0.7 & {11.9} $\pm$ {1.4} & {14.1} $\pm$ {1.1} \\
        FruitBot & 29.9 $\pm$ 0.5 & 29.1 $\pm$ 1.3 & 28.3 $\pm$ 1.3 & 29.8 $\pm$ 1.1 & {31.9} $\pm$ 0.4 & {31.9} $\pm$ 0.4 & 31.1 $\pm$ 0.8 \\
        Heist & 7.6 $\pm$ 0.6 & 7.2 $\pm$ 0.6 & 8.0 $\pm$ 0.5 & 4.9 $\pm$ 0.8 & 7.4 $\pm$ 0.6 & {8.6} $\pm$ {0.4} & {9.4} $\pm$ {0.4}\\
        Jumper & 8.7 $\pm$ 0.3 & 8.4 $\pm$ 0.5 & 8.6 $\pm$ 0.3 & 8.8 $\pm$ 0.3 & {9.1} $\pm$ {0.3} & 8.6 $\pm$ 0.4 & 8.9 $\pm$ 0.3 \\
        Leaper & 5.8 $\pm$ 1.5 & 3.6 $\pm$ 1.1 & 7.0 $\pm$ 0.6 & {8.8} $\pm$ {1.0} & 8.2 $\pm$ 2.9 & 8.4 $\pm$ 2.3 & 8.4 $\pm$ 2.3 \\
        Maze & 9.3 $\pm$ 0.3 & 8.5 $\pm$ 0.5 & 9.2 $\pm$ 0.4 & 5.7 $\pm$ 0.4 & 9.5 $\pm$ 0.3 & {9.6} $\pm$ {0.2} & {9.7} $\pm$ {0.2}\\
        Miner & 12.8 $\pm$ 0.2 & 12.4 $\pm$ 0.3 & 11.4 $\pm$ 0.6 & 4.7 $\pm$ 0.7 & 11.6 $\pm$ 0.4 & 12.7 $\pm$ 0.1 & {12.8} $\pm$ {0.1} \\
        Ninja & 7.8 $\pm$ 0.4 & 7.5 $\pm$ 1.0 & 8.2 $\pm$ 0.4 & 9.0 $\pm$ 0.2 & {9.8} $\pm$ {0.2} & 9.6 $\pm$ 0.1 & 9.6 $\pm$ 0.2 \\
        Plunder & 6.2 $\pm$ 0.7 & 9.2 $\pm$ 1.3 & 11.0 $\pm$ 1.1 & 5.9 $\pm$ 0.6 & 14.5 $\pm$ 2.1 & {15.7} $\pm$ {1.8} & {16.0} $\pm$ {2.3} \\
        StarPilot & 29.9 $\pm$ 3.6 & 30.2 $\pm$ 2.4 & 26.3 $\pm$ 3.2 & 39.0 $\pm$ 2.5 & 48.4 $\pm$ 3.4 & {50.5} $\pm$ {1.8} & {50.8} $\pm$ {4.0} \\
        \midrule
       PPO-norm score (\%) & 100.0 $\pm$ 2.5 & 102.6 $\pm$ 3.4 & 107.9 $\pm$ 3.5 & 104.4 $\pm$ 4.1 & 136.7 $\pm$ 4.4 & \textbf{148.9} $\pm$ \textbf{4.3} & \textbf{152.7} $\pm$ \textbf{4.4} \\
        \bottomrule
    \end{tabular}
    }
\end{table}

\section{Ablation Studies}

\subsection{Delayed Critic Update}

To disentangle the effect of the delayed critic update on learning generalizable representations for the policy network, we introduce a variant of DCPG that employs an additional value network $V_\phi$ with a separate encoder, namely Separate DCPG. The separate value network $V_\phi$ is trained in the same way as PPG without any delayed update and used only for policy optimization. The original value network $V_\theta$ is trained with the delayed update and used only for learning representation for the policy network.

\Cref{alg:dcpg_sep} describes the detailed procedure of Separate DCPG (differences with DCPG are marked in cyan). Note that we compute two bootstrapped value function targets $\hat{R}_{t, \theta}$ and $\hat{R}_{t, \phi}$ for each state $s_t$ using the original value network $V_\theta$ and the separate value network $V_\phi$, respectively. Each value function target is stored in a buffer $\mathcal{B}$ and used to train the corresponding value network during the auxiliary phase.
 
\begin{algorithm}[h]
	\small
	\caption{\cyan{Separate} Delayed-Critic Policy Gradient (\cyan{Separate} DCPG)}
	\begin{algorithmic}[1]
	\Require Policy network $\pi_\theta$, value network $V_\theta$, \cyan{separate value network $V_\phi$}
    	\For{phase = $1, 2, \ldots $}
    	\State Initialize buffer $\mathcal{B}$
    	\For{iter = $1, 2, \ldots, N_\pi$} \Comment{Policy phase}
    	\State Sample trajectories $\tau$ using $\pi_\theta$
	\State Compute value function targets $\hat{R}_{t, \theta}$ \cyan{and $\hat{R}_{t, \phi}$} for each state $s_t \in \tau$
    	\For{epoch = $1, 2, \ldots, E_\pi$}
    	\State Optimize policy objective $J_\pi(\theta)$ and value regularizer $C_V(\theta)$ with $\tau$
	\State \cyan{Optimize value objective $J_V(\phi)$ with $\tau$}
    	\EndFor
	\State Add $(s_t, \hat{R}_{t, \theta}, \cyan{\hat{R}_{t, \phi}})$ to $\mathcal{B}$
    	\EndFor
	\For{iter = $1, 2, \ldots, E_\mathrm{aux}$} \Comment{Auxiliary phase}
	\State Optimize value objective $J_V(\theta)$ and policy regularizer $C_\pi(\theta)$ with $\mathcal{B}$
	\State \cyan{Optimize value objective $J_V(\phi)$ with $\mathcal{B}$}
	\EndFor
    	\EndFor
	\end{algorithmic}
	\label{alg:dcpg_sep}
\end{algorithm}

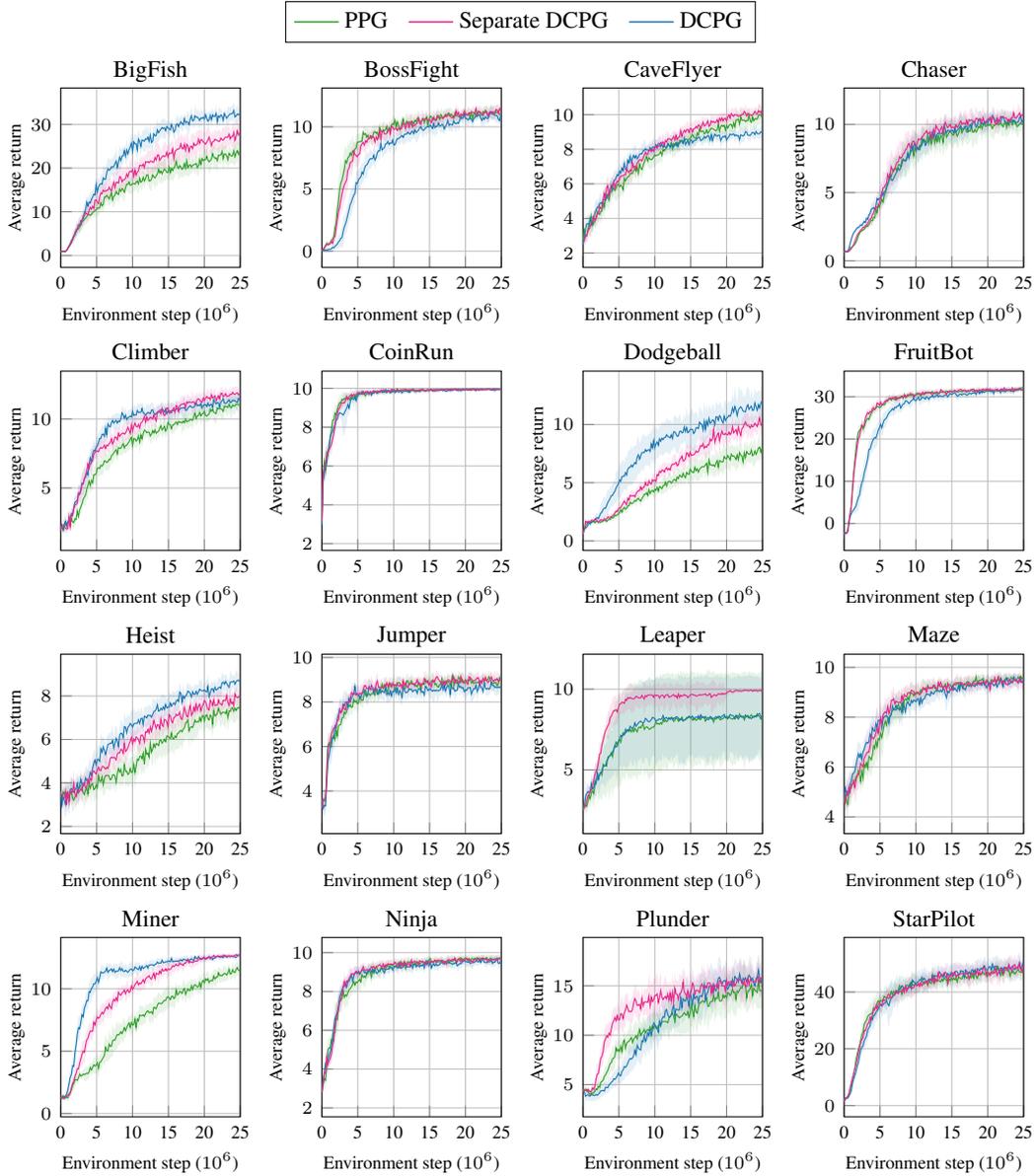
\begin{figure}[H]
    \centering
    \begin{tikzpicture}
        \begin{groupplot}[
		group style = {group name=procgen, group size = 4 by 4, horizontal sep=1.1cm, vertical sep=1.4cm},
		width=4.0cm,
		height=4.0 cm,
		grid=major,
		xmin=0,
		xmax=2.5e7,
		scaled x ticks = false,
		tick pos = left,
		tick label style={font=\scriptsize},
		xtick={0, 5e6, 10e6, 15e6, 20e6, 25e6},
		xticklabels={0, 5, 10, 15, 20, 25},
		xlabel={Environment step ($10^6$)},
		ylabel={Average return},
		label style={font=\scriptsize},
		ylabel style={at={(-0.15,0.5)}},
		xlabel style={at={(0.5,-0.15)}},
		title style={font=\small,anchor=north,yshift=8pt},
		legend style={font=\small,legend columns=3, /tikz/every even column/.append style={column sep=0.2cm}},
		legend to name=exp,
        ]
		\pgfplotsinvokeforeach{BigFish,BossFight,CaveFlyer,Chaser,Climber,CoinRun,Dodgeball,FruitBot,Heist,Jumper,Leaper,Maze,Miner,Ninja,Plunder,StarPilot}{
		    \nextgroupplot[
		        title=#1,        
		    ]
		\addlegendimage{index of colormap=3 of Paired,line width=0.2mm}
    		\addlegendentry{PPG}
		\addlegendimage{index of colormap=3 of Dark2,line width=0.2mm}
    		\addlegendentry{Separate DCPG}
		\addlegendimage{index of colormap=1 of Paired,line width=0.2mm}
    		\addlegendentry{DCPG}

    		\addplot+[index of colormap=3 of Paired,line width=0.15mm] table [x=num_steps, y=train_episode_rewards_mean, col sep=comma]{data/progress-#1-ppg.csv};
    		\addplot+[index of colormap=1 of Paired,line width=0.15mm] table [x=num_steps, y=train_episode_rewards_mean, col sep=comma]{data/progress-#1-appg.csv};
    		\addplot+[index of colormap=3 of Dark2,line width=0.15mm] table [x=num_steps, y=train_episode_rewards_mean, col sep=comma]{data/progress-#1-appg_sep.csv};
    		\addplot[name path=plus,draw=none] table [x=num_steps, y=train_episode_rewards_mean_plus_std, col sep=comma]{data/progress-#1-ppg.csv};
    		\addplot[name path=minus,draw=none] table [x=num_steps, y=train_episode_rewards_mean_minus_std, col sep=comma]{data/progress-#1-ppg.csv};
    		\addplot[index of colormap=3 of Paired,fill opacity=0.12] fill between[of=plus and minus];
    		\addplot[name path=plus,draw=none] table [x=num_steps, y=train_episode_rewards_mean_plus_std, col sep=comma]{data/progress-#1-appg.csv};
    		\addplot[name path=minus,draw=none] table [x=num_steps, y=train_episode_rewards_mean_minus_std, col sep=comma]{data/progress-#1-appg.csv};
    		\addplot[index of colormap=1 of Paired,fill opacity=0.12] fill between[of=plus and minus];
    		\addplot[name path=plus,draw=none] table [x=num_steps, y=train_episode_rewards_mean_plus_std, col sep=comma]{data/progress-#1-appg_sep.csv};
    		\addplot[name path=minus,draw=none] table [x=num_steps, y=train_episode_rewards_mean_minus_std, col sep=comma]{data/progress-#1-appg_sep.csv};
    		\addplot[index of colormap=3 of Dark2,fill opacity=0.12] fill between[of=plus and minus];
    		}
        \end{groupplot}
	\node[above] at (current bounding box.north) {\pgfplotslegendfromname{exp}};
    \end{tikzpicture}
    \caption{Training performance curves of PPG, Separate DCPG, and DCPG on all 16 procgen games. Each agent is trained on 200 training levels for 25M environment steps and evaluated on the same training levels. The mean and standard deviation are computed over 10 different runs.}
    \label{fig:procgen_train_ablation_full}
\end{figure}

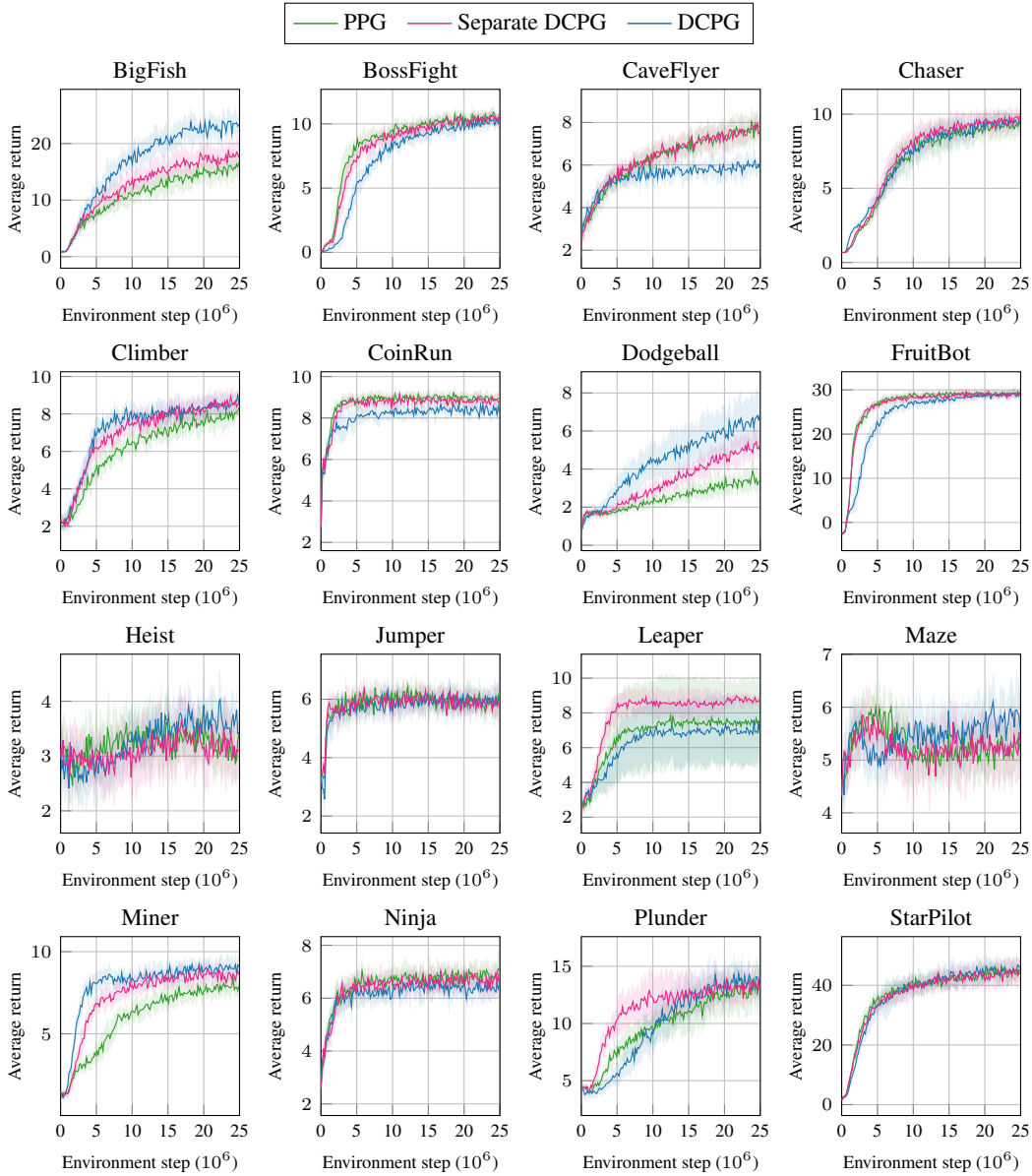
\begin{figure}[H]
    \centering
    \begin{tikzpicture}
        \begin{groupplot}[
		group style = {group name=procgen, group size = 4 by 4, horizontal sep=1.1cm, vertical sep=1.4cm},
		width=4.0cm,
		height=4.0 cm,
		grid=major,
		xmin=0,
		xmax=2.5e7,
		scaled x ticks = false,
		tick pos = left,
		tick label style={font=\scriptsize},
		xtick={0, 5e6, 10e6, 15e6, 20e6, 25e6},
		xticklabels={0, 5, 10, 15, 20, 25},
		xlabel={Environment step ($10^6$)},
		ylabel={Average return},
		label style={font=\scriptsize},
		ylabel style={at={(-0.15,0.5)}},
		xlabel style={at={(0.5,-0.15)}},
		title style={font=\small,anchor=north,yshift=8pt},
		legend style={font=\small,legend columns=3, /tikz/every even column/.append style={column sep=0.2cm}},
		legend to name=exp,
        ]
		\pgfplotsinvokeforeach{BigFish,BossFight,CaveFlyer,Chaser,Climber,CoinRun,Dodgeball,FruitBot,Heist,Jumper,Leaper,Maze,Miner,Ninja,Plunder,StarPilot}{
		    \nextgroupplot[
		        title=#1,        
		    ]
		\addlegendimage{index of colormap=3 of Paired,line width=0.2mm}
    		\addlegendentry{PPG}
		\addlegendimage{index of colormap=3 of Dark2,line width=0.2mm}
    		\addlegendentry{Separate DCPG}
		\addlegendimage{index of colormap=1 of Paired,line width=0.2mm}
    		\addlegendentry{DCPG}

    		\addplot+[index of colormap=3 of Paired,line width=0.15mm] table [x=num_steps, y=test_episode_rewards_mean, col sep=comma]{data/progress-#1-ppg.csv};
    		\addplot+[index of colormap=1 of Paired,line width=0.15mm] table [x=num_steps, y=test_episode_rewards_mean, col sep=comma]{data/progress-#1-appg.csv};
    		\addplot+[index of colormap=3 of Dark2,line width=0.15mm] table [x=num_steps, y=test_episode_rewards_mean, col sep=comma]{data/progress-#1-appg_sep.csv};
    		\addplot[name path=plus,draw=none] table [x=num_steps, y=test_episode_rewards_mean_plus_std, col sep=comma]{data/progress-#1-ppg.csv};
    		\addplot[name path=minus,draw=none] table [x=num_steps, y=test_episode_rewards_mean_minus_std, col sep=comma]{data/progress-#1-ppg.csv};
    		\addplot[index of colormap=3 of Paired,fill opacity=0.12] fill between[of=plus and minus];
    		\addplot[name path=plus,draw=none] table [x=num_steps, y=test_episode_rewards_mean_plus_std, col sep=comma]{data/progress-#1-appg.csv};
    		\addplot[name path=minus,draw=none] table [x=num_steps, y=test_episode_rewards_mean_minus_std, col sep=comma]{data/progress-#1-appg.csv};
    		\addplot[index of colormap=1 of Paired,fill opacity=0.12] fill between[of=plus and minus];
    		\addplot[name path=plus,draw=none] table [x=num_steps, y=test_episode_rewards_mean_plus_std, col sep=comma]{data/progress-#1-appg_sep.csv};
    		\addplot[name path=minus,draw=none] table [x=num_steps, y=test_episode_rewards_mean_minus_std, col sep=comma]{data/progress-#1-appg_sep.csv};
    		\addplot[index of colormap=3 of Dark2,fill opacity=0.12] fill between[of=plus and minus];
    		}
        \end{groupplot}
	\node[above] at (current bounding box.north) {\pgfplotslegendfromname{exp}};
    \end{tikzpicture}
    \caption{Test performance curves of PPG, Separate DCPG, and DCPG on all 16 procgen games. Each agent is trained on 200 training levels for 25M environment steps and evaluated on 100 unseen test levels. The mean and standard deviation are computed over 10 different runs.}
    \label{fig:procgen_test_ablation_full}
\end{figure}

\Cref{fig:procgen_train_ablation_full,fig:procgen_test_ablation_full} demonstrate the training and test performance curves of PPG, Separate DCPG, and DCPG. We find that Separate DCPG achieves better or comparable performance in most games. This implies that the value network with delayed updates can provide better representations that generalize well to both the training and test environments than those without delayed updates. We also provide the PPO-normalized training and test scores of PPG, Separate DCPG, and DCPG in \Cref{tab:procgen_dcpg_sep}.

\newpage

\begin{table}[H]
	\centering
	\caption{PPO-normalized training and test scores of PPG, Separate DCPG, and DCPG on all 16 Procgen games. Each agent is trained on 200 training levels for 25M environment steps. The mean and standard deviation are computed over 10 different runs.}
    	\begin{tabular}{l|>{\raggedleft}p{2cm}>{\raggedleft}p{2.4cm}>{\raggedleft}p{2cm}}
	\toprule
	 & PPG & Separate DCPG & DCPG \\
	\midrule
	PPO-norm train score (\%) & 136.7 $\pm$ 4.4 & 147.5 $\pm$ 1.9 & \textbf{152.7} $\pm$ \textbf{4.4} \\
	PPO-norm test score (\%) & 160.3 $\pm$ 6.3 & 174.4 $\pm$ 4.9 & \textbf{184.5} $\pm$ \textbf{5.2} \\
	\bottomrule
	\end{tabular}
    	\label{tab:procgen_dcpg_sep}
\end{table}

\newpage

\subsection{Dynamics Learning}

To demonstrate the effectiveness of learning forward and inverse dynamics with a single discriminator, we conduct experiments for DCPG with forward dynamics learning (DCPG+F), inverse dynamics learning (DCPG+I), and forward and inverse dynamics learning using two separate discriminators (DCPG+FI). The dynamics objective $J_f$ of each algorithm is defined as follows:
\begin{align*}
	J_f(\theta) &= \mathbb{E}_{s_t, a_t, s_{t+1} \sim \mathcal{B}} \left[ \log \left(f_\theta \left(s_t, a_t, {s}_{t+1} \right)\right) + \log \left(1 - f_\theta \left( s_t, a_t, \hat{s}_{t+1} \right) \right) \right] \tag{DCPG+F} \\
	J_f(\theta) &= \mathbb{E}_{s_t, a_t, s_{t+1} \sim \mathcal{B}} \left[ \log \left(f_\theta \left( s_t, a_t, {s}_{t+1} \right)\right) + \log \left(1 - f_\theta \left( s_t, \hat{a}_t, s_{t+1} \right) \right) \right] \tag{DCPG+I} \\
	J_f(\theta) &=  \mathbb{E}_{s_t, a_t, s_{t+1} \sim \mathcal{B}} \left[ \log \left(f_\theta \left( s_t, a_t, {s}_{t+1} \right)\right) + \log \left(1 - f_\theta \left( s_t, a_t, \hat{s}_{t+1} \right) \right) \right] \\
	&\quad+ \eta~ \mathbb{E}_{s_t, a_t, s_{t+1} \sim \mathcal{B}} \left[ \log \left(\tilde{f}_\theta \left( s_t, a_t, {s}_{t+1} \right)\right) + \log \left(1 - \tilde{f}_\theta \left( s_t, \hat{a}_t, s_{t+1} \right) \right) \right]. \tag{DCPG+FI}
\end{align*}
Note that we use an additional discriminator $\tilde{f}$ for inverse dynamics in DCPG+FI. For each algorithm, we train agents using 200 training levels for 25M environment steps on 16 Procgen games. We set the dynamics objective coefficient to $\beta_f = 1.0$ without any hyperparameter tuning. For DCPG+FI, we sweep the inverse dynamics coefficient within a range of $\eta \in \{ 0.5, 1.0 \}$ and choose $\eta=1.0$. \Cref{tab:procgen_ssl_full} shows the PPO-normalized training and test scores of each algorithm.

\begin{table}[h]
	\centering
	\caption{PPO-normalized training and test scores of DCPG and DCPG with dynamics learning on all 16 Procgen games. Each agent is trained on 200 training levels for 25M environment steps. The mean and standard deviation are computed over 10 different runs.}
    	\adjustbox{max width=\textwidth}{
	\begin{tabular}{l|rrrrr}
	\toprule
	 & DCPG & DCPG+F & DCPG+I & DCPG+FI & DDCPG \\
	\midrule
	PPO-norm train score (\%) & 148.9 $\pm$ 4.3 & 150.6 $\pm$ 3.1 & 149.1 $\pm$ 3.4 & 150.5 $\pm$ 2.5 & \textbf{152.7} $\pm$ \textbf{4.4} \\
	PPO-norm test score (\%) & 184.5 $\pm$ 5.2 & 195.9 $\pm$ 7.7 & 185.5 $\pm$ 6.5 & 194.6 $\pm$ 6.2 & \textbf{202.2} $\pm$ \textbf{10.2} \\
	\bottomrule
	\end{tabular}}
    	\label{tab:procgen_ssl_full}
\end{table}

Our intuition of jointly learning forward and inverse dynamics using a single discriminator is that naively learning the forward dynamics will discard action information and capture only the proximity of two consecutive states in the latent space, not the dynamics. Also, additional training of the inverse dynamics with a separate discriminator cannot completely resolve this problem.

To validate this, we train three types of DCPG agents with dynamics learning, DCPG+F, DCPG+FI, and DDCPG on BigFish and count the number of OOD actions that the forward dynamics discriminator determines to be valid given a transition $(s_t, a_t, s_{t+1})$ from the training environments, \ie, $\sum_{\hat{a}_t \neq a_t} \mathds{1}[f(s_t, \hat{a}_t, s_{t+1}) > 0.5]$. Note that there are 9 different actions (8 directional moves and 1 do nothing) in BigFish. \Cref{tab:procgen_ssl_count} shows that the discriminators of DCPG+F and DCPG+FI determine multiple OOD actions to be valid, implying that it does not fully utilize the action information.

\begin{table}[h]
	\centering
	\caption{The number of OOD actions classified as valid for DCPG+F, DCPG+FI, and DDCPG on BigFish. Each agent is trained on 200 training levels for 25M environment steps.}
    	\begin{tabular}{l|rrr}
	\toprule
	 & DCPG+F & DCPG+FI & DDCPG \\
	\midrule
	\# OOD actions ($\downarrow$) & 7.05 $\pm$ 1.47 & 2.49 $\pm$ 1.38 & \textbf{0.74} $\pm$ \textbf{0.75} \\
	\bottomrule
	\end{tabular}
    	\label{tab:procgen_ssl_count}
\end{table}

Furthermore, we train PPG agents with our proposed dynamics learning (named DPPG) to check whether the effect of the delayed critic update and the dynamics learning are complementary. \Cref{tab:procgen_ssl_ppg} shows that dynamics learning is also helpful to PPG, while the extent of performance improvement is smaller than DCPG. It implies that the effects of the delayed value update and the dynamics learning are synergistic.

\begin{table}[h]
	\centering
	\caption{PPO-normalized test scores of PPG, DPPG, DCPG, and DDCPG on all 16 Procgen games. Each agent is trained on 200 training levels for 25M environment steps. The mean and standard deviation are computed over 10 different runs.}
    	\begin{tabular}{l|rrrr}
	\toprule
	 & PPG & DPPG & DCPG & DDCPG \\
	\midrule
	PPO-norm score (\%) & 160.3 $\pm$ 6.3 & 171.7 $\pm$ 4.9 & 184.5 $\pm$ 5.2 & 202.2 $\pm$ 10.2  \\
	\bottomrule
	\end{tabular}
    	\label{tab:procgen_ssl_ppg}
\end{table}

\section{Evaluation using RLiable \citep{agarwal2021deep}}

\begin{figure}[t]
	\includegraphics[width=\textwidth]{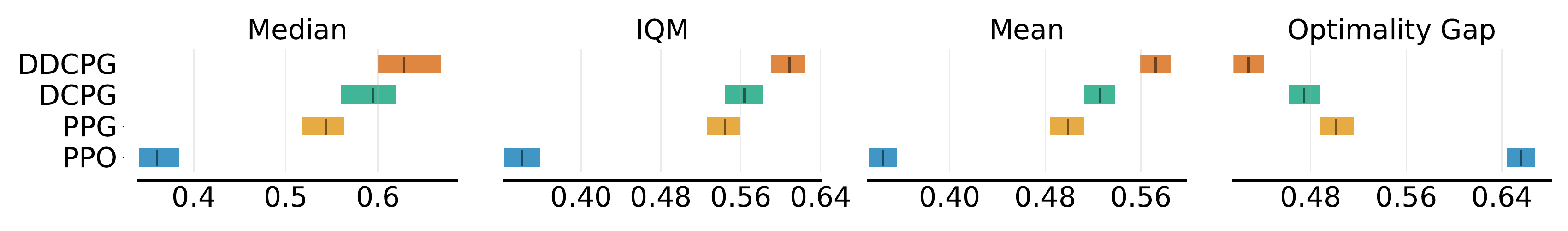}
	\caption{Median/IQM/Mean of Min-Max normalized scores with 95\% confidence intervals for PPO, PPG, DCPG, and DDCPG on all 16 Procgen games. Each statistic is computed over 10 seeds.}
	\label{fig:iqm}
\end{figure}

\begin{figure}[t]
	\centering
	\includegraphics[width=0.7\textwidth]{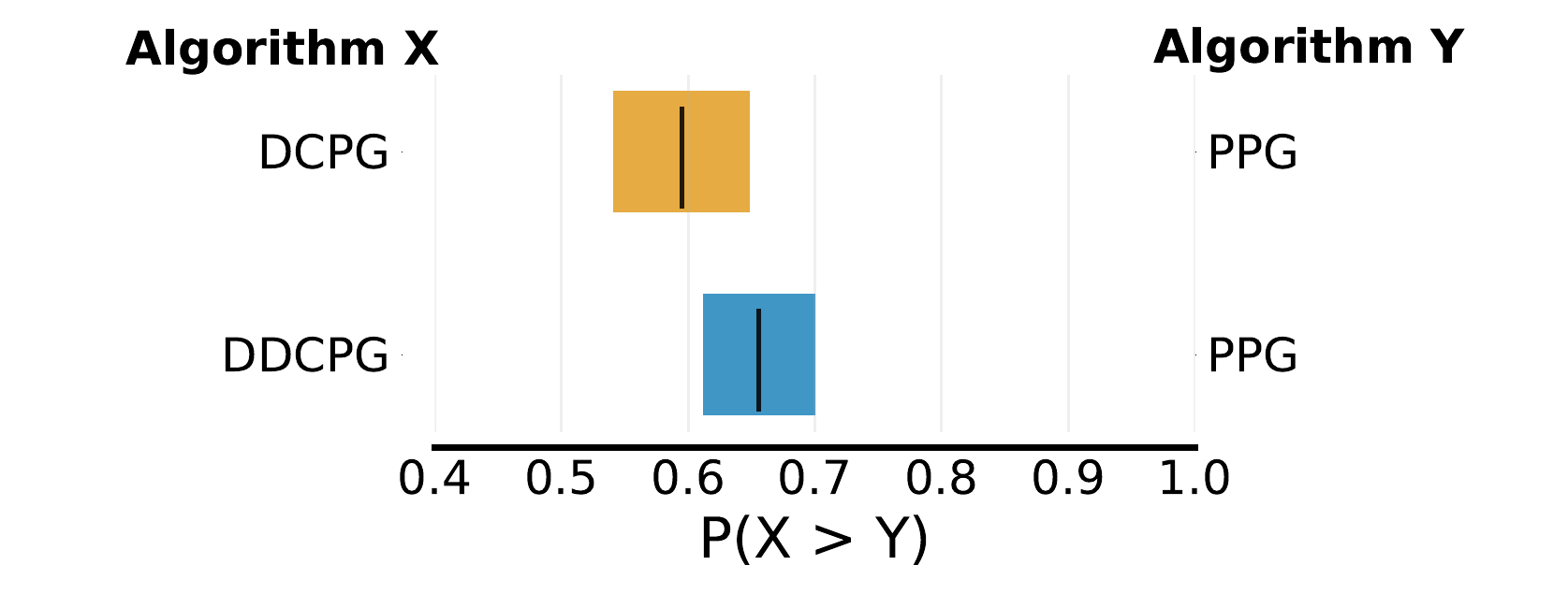}
	\caption{Probability of improvement of DCPG and DDCPG compared to PPG on all 16 Procgen games. Each statistic is computed over 10 seeds.}
	\label{fig:improvement}
\end{figure}

We also evaluated our experimental results by normalizing the average returns based on the possible minimum and maximum scores for each game and analyzing the min-max normalized scores using the RLiable library\footnote{\url{https://github.com/google-research/rliable}}. \Cref{fig:iqm} reports the Median, IQM, and Mean scores of PPO, PPG, and our methods, showing that the performance improvements of our methods are statistically significant in terms of all evaluation metrics. \Cref{fig:improvement} describes the probability of improvement plots comparing our methods to PPG, showing that our methods are likely to improve upon PPG.

\end{document}